\documentclass{article}

\usepackage[preprint,nonatbib]{neurips_2026}

\usepackage[utf8]{inputenc}
\usepackage[T1]{fontenc}
\usepackage{hyperref}
\usepackage{url}
\usepackage{booktabs}
\usepackage{amsfonts}
\usepackage{amsmath}
\usepackage{microtype}
\usepackage{xcolor}
\usepackage{colortbl}
\definecolor{rowhighlight}{rgb}{0.90,0.90,0.97}
\usepackage{graphicx}
\usepackage{multirow}
\usepackage[numbers,sort&compress]{natbib}
\usepackage{caption}
\usepackage{titlesec}
\titlespacing*{\paragraph}{0pt}{4pt}{1em}

\title{DecisionBench: A Benchmark for Emergent Delegation in Long-Horizon Agentic Workflows}

\author{%
  \textbf{Yuxuan Gao}$^{1,2*}$ \quad
  \textbf{Megan Wang}$^{1,3*}$ \quad
  \textbf{Yi Ling Yu}$^{1,2*}$ \quad
  \textbf{Zijian Carl Ma}$^{4}$ \quad
  \textbf{Ao Qu}$^{5}$ \\[6pt]
  $^{1}$OpenMesh AI \quad
  $^{2}$University of Pennsylvania \quad
  $^{3}$Columbia University \\[2pt]
  $^{4}$Stanford University \quad
  $^{5}$MIT \\[2pt]
  $*$ Equal contribution (alphabetical order) \\[4pt]
  \textbf{Code and data:} \url{https://huggingface.co/decisionbench}
}

\begin{document}
\maketitle

\begin{abstract}
We introduce \textbf{DecisionBench}, a benchmark substrate for
emergent delegation in long-horizon agentic workflows. The
substrate fixes a task suite (GAIA, $\tau$-bench, BFCL multi-turn),
a peer-model pool (11 models, 7 vendor families), a delegation
interface (\texttt{call\_model} plus an optional
\texttt{read\_profile} channel), a deterministic skill-annotation
layer, and a multi-axis metric suite covering quality, cost,
latency, delegation rate, routing fidelity-at-$k$, vendor
self-preference, and a counterfactual-delegation ceiling. The
substrate is agnostic to how peer information is generated or
delivered, so learned routers, richer peer memories, adaptive
profile construction, and multi-step delegation can all be
evaluated against it. We characterize the substrate with a
five-condition reference sweep on the full pool ($n{=}23{,}375$
task instances). Three benchmark-level findings emerge: (i)~mean
end-task quality is statistically indistinguishable across the four
awareness conditions ($|\beta|{\leq}0.010$, $p{\geq}0.21$), so
quality-only evaluation would miss the orchestration signal;
(ii)~routing fidelity-at-$1$ ranges from $7.5\%$ to $29.5\%$ across
conditions at near-equal mean quality, with delivery channel
(on-demand tool vs.\ preloaded description) dominating description
content; (iii)~a counterfactual ceiling places perfect delegation
$15$--$31$ percentage points above measured performance on every
suite, locating large unrealized headroom for future orchestration
methods. We release the substrate, annotation layer, reference
intervention suite, analysis pipeline, and 220 per-condition run
archives.
\end{abstract}

\section{Introduction}
\label{sec:intro}

Large language model (LLM) agents are increasingly deployed for tasks
that span minutes to hours of tool
use~\citep{react,agentsurvey},
with the dominant cost shifting from per-token quality to compute
spent on sub-tasks where a smaller or differently-specialized peer
model would have sufficed~\citep{routellm,frugalgpt,routerbench}.
Routing a sub-task to a peer rather than solving it locally is
therefore a Pareto-frontier shifter alongside raw model capability.
A practitioner deploying an agentic system today must decide whether
to delegate, to whom, and what (if anything) to tell the orchestrator
about its peers. These decisions are typically made by intuition
because no benchmark has measured them in isolation.

Existing agentic benchmarks measure single-agent capability on a fixed
task --- GAIA~\citep{gaia}, $\tau$-bench~\citep{taubench},
BFCL~\citep{bfcl} and others (\S\ref{sec:related}). Multi-agent systems
work~\citep{autogen,metagpt} typically uses hand-coded role assignment
rather than emergent delegation. Cost-aware routing
benchmarks~\citep{routellm,routerbench} treat routing as a learned
external policy rather than a behavior the orchestrator must exhibit
on its own. SkillsBench~\citep{skillsbench} measures skill scaffolds
attached to a single agent. None of these directly measure whether an
agent delegates well between peer models in the course of solving a
long-horizon task, or what process-level metrics would reveal about
how those delegation decisions are made.

We introduce \textbf{DecisionBench}, a benchmark substrate for
emergent delegation in long-horizon agentic workflows. The substrate
fixes (1) a task suite (GAIA, $\tau$-bench, BFCL multi-turn, with
deterministic Stage-1/Stage-2 splits), (2) a peer-model pool (11
models from 7 vendor families pinned to a freeze date), (3) a
delegation interface (a \texttt{call\_model} tool, optionally paired
with a \texttt{read\_profile} channel for peer-information
interventions), (4) an annotation layer (a frozen 7-skill taxonomy
and a deterministic step tagger, released as evaluation machinery
rather than a method we propose), and (5) a metric suite covering
end-task quality, cost, latency, delegation rate, routing
fidelity-at-$k$, vendor self-preference, and a
counterfactual-delegation ceiling. The substrate is agnostic to how
peer information is generated or delivered; methods that produce
learned routers, richer peer memories, adaptive profile construction,
or multi-step delegation can all be evaluated against it.

To demonstrate how DecisionBench evaluates peer-awareness
interventions, we instantiate four reference conditions over the
released annotation layer, against a no-information baseline:
(\texttt{blind}) only \texttt{call\_model}, no peer description;
(\texttt{aware-c1}/\texttt{c2}/\texttt{c3}) preload a peer
description constructed three different ways --- a curated rubric, deterministic
statistics from Stage-1 traces, or dual out-of-pool LLM-judge
summaries --- and expose \texttt{read\_profile}; and
(\texttt{aware-tool-only}) the same tools as the aware variants but
with the preloaded description suppressed, a one-variable ablation
that isolates delivery channel from description content. These
reference interventions are not part of the benchmark definition;
they are the baselines we use to demonstrate the substrate and to
seed comparisons for future methods.

\paragraph{Findings.}
Across the 5-condition reference sweep on the 11-model pool
($11{\times}3{\times}5{=}165$ cells, $n{=}23{,}375$ tasks), three
benchmark-level patterns emerge.
(1)~\emph{End-task quality is mostly flat across awareness
conditions} ($|\beta|{\leq}0.010$, $p{\geq}0.21$ in a mixed-effects
fit), so quality-only evaluation would miss the orchestration signal
entirely.
(2)~\emph{Delegation fidelity moves cleanly}: on-demand tool access
(\texttt{aware-tool-only}) more than doubles routing
precision-at-$1$ over \texttt{blind} ($14.2\%{\to}29.5\%$) at
quality parity and lower mean cost; preloaded variants capture less
than half of the gain (C2 $20.8\%$, C3 $15.5\%$; C1 $7.5\%$, below
blind). The dissociation between fidelity and quality is itself the
methodological payoff of measuring process-level metrics, not just
outcomes.
(3)~A counterfactual-delegation ceiling places perfect single-step
delegation 15--31~pp above measured performance on every suite,
locating substantial unrealized headroom for future orchestration
methods. We additionally document cross-vendor self-preference in
delegation behavior at $1.5$--$3.7\times$ chance.

\paragraph{Contributions.}
(1)~\textbf{The DecisionBench benchmark} for emergent delegation in
long-horizon agentic workflows: a fixed task suite, peer-model
pool, delegation interface, annotation layer (frozen 7-skill
taxonomy plus deterministic step tagger), and a multi-axis metric
suite (quality, cost, latency, delegation rate, fidelity-at-$k$,
vendor self-preference, counterfactual ceiling).
(2)~\textbf{A reference intervention suite} for peer-awareness on
the benchmark: three profile-card content variants (curated rubric,
deterministic statistics, dual out-of-pool LLM-judge summary) and
a delivery-channel ablation (\texttt{aware-tool-only}) that
isolates information content from delivery mechanism. These are
baselines on DecisionBench, not part of the benchmark definition.
(3)~\textbf{An empirical characterization} of the substrate using
the reference suite: end-task quality is statistically flat across
awareness conditions, delivery channel dominates description
content, and the perfect-delegation ceiling exceeds measured
performance by 15--31~pp on every suite --- patterns visible only
through the process-level metrics the benchmark instruments.
(4)~\textbf{Documented cross-vendor self-preference} in delegation
behavior, the orchestration-tool analogue of LLM-as-judge
biases~\citep{zheng2023judging,selfpreference,positionbias,lengthbias}.

\begin{figure}[!ht]
\centering
\includegraphics[width=\textwidth]{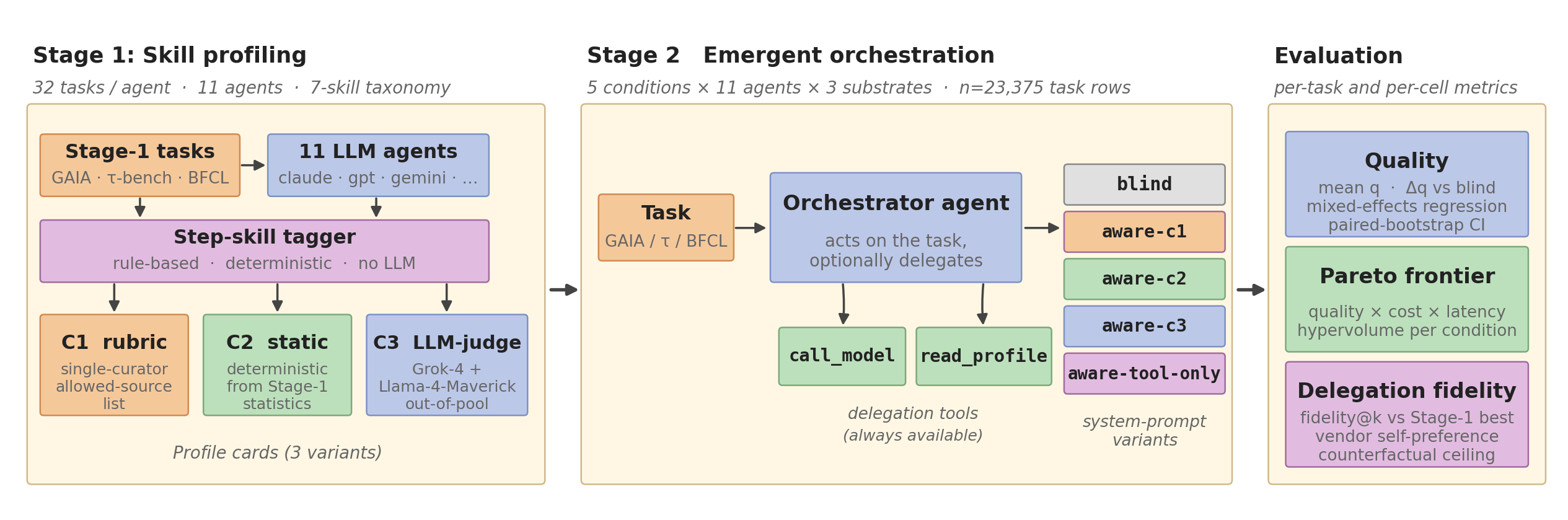}
\caption{\textbf{DecisionBench overview.} \emph{Left (substrate, \S\ref{sec:design}).}
The benchmark fixes a task suite (GAIA, $\tau$-bench, BFCL
multi-turn), a peer-model pool (11 models, 7 vendor families), a
delegation interface (\texttt{call\_model}, plus an optional
\texttt{read\_profile} channel), an annotation layer (frozen 7-skill
taxonomy and deterministic step tagger;
App.~\ref{app:taxonomy}), and a metric suite (quality, cost,
latency, delegation rate, fidelity-at-$k$, self-preference,
counterfactual ceiling).
\emph{Right (reference interventions, \S\ref{sec:conditions}).}
To demonstrate the substrate, peer trajectories on the Stage-1 split
are rendered into three profile-card variants (\texttt{C1} rubric,
\texttt{C2} deterministic statistics, \texttt{C3} dual LLM-judge);
each is used as a preloaded description in an \texttt{aware-}
condition. A delivery-channel ablation
(\texttt{aware-tool-only}) exposes \texttt{read\_profile} without
preloading any description, isolating delivery from content. These
five conditions seed comparisons for future orchestration methods
but are not part of the benchmark definition.}
\label{fig:overview}
\end{figure}

\section{Related Work}
\label{sec:related}

\paragraph{Agentic benchmarks.}
GAIA~\citep{gaia} introduced a general-assistant benchmark framed against
the single-shot evaluation limitation; $\tau$-bench~\citep{taubench}
tests multi-turn state tracking under domain policy~\citep{mteval,mtbench};
SWE-Bench~\citep{swebench} and SWE-Bench Pro~\citep{swebenchpro} cover
repository-level coding; BFCL~\citep{bfcl} and
ToolLLM~\citep{toolllm} measure function-calling correctness;
WebArena~\citep{webarena}, Mind2Web~\citep{mind2web}, and
WebShop~\citep{webshop} cover web environments;
ALFWorld~\citep{alfworld} and OSWorld~\citep{osworld} cover embodied /
desktop-OS settings; AgentBench~\citep{agentbench} aggregates many;
MLE-bench~\citep{mlebench} and Cybench~\citep{cybench} target ML
engineering and cybersecurity. We use GAIA, $\tau$-bench, and BFCL as
heterogeneous quality probes spanning open-ended retrieval, dialogue,
and structured tool calls.

\paragraph{Skill-aware agent design.}
SkillsBench~\citep{skillsbench} measures the contribution of skill
scaffolds to agent configurations such as Claude Code and Codex CLI;
related lines extend single-agent capability through chained
reasoning~\citep{cot,tot,planandsolve,decomposed},
self-feedback~\citep{reflexion,selfrefine}, open-ended
exploration~\citep{voyager,generativeagents}, constitution-style
behavior shaping~\citep{cai,instructgpt}, and long-context
processing~\citep{longbench,ruler}. DecisionBench differs in (1)
testing raw model APIs rather than product configurations and
(2) treating ``skill knowledge'' as peer capabilities surfaced to a
different agent that may delegate, letting us measure whether
structured peer knowledge improves cross-model delegation --- a
question single-agent skill scaffolds cannot address.

\paragraph{Cost-aware orchestration and routing.}
RouteLLM~\citep{routellm} and FrugalGPT~\citep{frugalgpt} treat
routing as a learned external policy;
HybridLLM~\citep{hybridllm}, AutoMix~\citep{automix},
EcoAssistant~\citep{ecoassistant}, and
RouterBench~\citep{routerbench} extend the policy line, while
mixture- and society-of-agents formulations~\citep{mixtureofagents,multiagentdebate,hugginggpt}
distribute work with hand-coded coordination. DecisionBench measures
emergent orchestration with no external policy and no hard-coded
roles --- the baseline before any router is trained.

\paragraph{LLM-as-judge for trace analysis.}
LLM judging~\citep{zheng2023judging,arenahard}
is now a standard substitute for human evaluation, with documented
self-preference~\citep{selfpreference}, position~\citep{positionbias},
and length~\citep{lengthbias} biases. DecisionBench's C3 profile variant
uses this technique for trace analysis (\S\ref{sec:profile-flavors});
we mitigate self-preference by using two judges from families
outside the agent pool and surfacing both judgments side-by-side
rather than averaging.

\section{DecisionBench: Benchmark Substrate}
\label{sec:design}

DecisionBench is a fixed substrate for measuring emergent delegation in
long-horizon agentic workflows. It comprises a task suite
(\S\ref{sec:tasks}), a peer-model pool (\S\ref{sec:pool}), a
delegation interface (\S\ref{sec:interface}), an annotation layer
(\S\ref{sec:taxonomy}) used as evaluation machinery rather than as a
proposed orchestration method, and a metric suite
(\S\ref{sec:metrics}). The substrate is agnostic to how peer
information is generated, summarized, or delivered; the reference
interventions we use to characterize it in \S\ref{sec:conditions} are
one family of methods among many it admits.

\subsection{Task suites and splits}
\label{sec:tasks}
DecisionBench draws long-horizon agentic tasks from three established
benchmarks, partitioned into a Stage-1 (profiling) split and a
Stage-2 (evaluation) split via a deterministic 20/80 stratified
sample (\texttt{seed=10}): GAIA~\citep{gaia} (general tool-use QA,
exact-match scoring; stratified by Level, 32~/~133 tasks within the
validation set since the public test-set leaderboard is
non-functional, see \S\ref{sec:limitations});
$\tau$-bench~\citep{taubench} (tool-agent-user dialogue,
pass\textsuperscript{$k$}; stratified by sub-domain into airline /
retail shards); BFCL v4 multi-turn~\citep{bfcl} (function-calling,
AST-match on per-turn calls, see \S\ref{sec:limitations}; stratified
by task family). The Stage-1 split is the basis for any
profile-building intervention (\S\ref{sec:conditions}); the Stage-2
split is the held-out evaluation set on which all reported numbers
are measured. ``Long-horizon'' here means multi-step trajectories
within a task, not multi-hour deployments. SWE-Bench Pro is excluded
from headline results (Docker harness unreliable in our regime) and
reported only as appendix material.

\subsection{Peer-model pool}
\label{sec:pool}
Eleven models across seven vendor families (OpenAI, Anthropic, Google
DeepMind, DeepSeek, Moonshot, Qwen, MiniMax), pinned to a 2026-04-29
freeze date and routed via OpenRouter~\citep{openrouter}. The pool is
intentionally heterogeneous in vendor, size class, and reasoning-mode
capability so that delegation is plausibly useful. Full per-model
list in Appendix~\ref{app:pool}.

\subsection{Delegation interface}
\label{sec:interface}
A DecisionBench orchestrator is an LLM agent running the Stage-2 task
loop with at least one tool, \texttt{call\_model(name, subtask,
budget\_usd)}, that delegates a sub-task to a named peer in the pool.
The peer receives only the orchestrator's \texttt{subtask} string
plus its own system prompt; it does not see earlier turns. The peer's
return value is the orchestrator's tool result, after which the
orchestrator continues the task. A delegation cap of 10 peer calls
per task bounds runaway loops; no other constraint is imposed.

The benchmark also defines an optional second tool,
\texttt{read\_profile(model)}, which returns a structured description
of the named peer. This tool is the channel through which
peer-information interventions are delivered to the orchestrator.
A method that does not consume peer information at all (e.g., the
\texttt{blind} baseline in \S\ref{sec:conditions}) simply omits this
tool. The benchmark does not specify what \texttt{read\_profile}
returns; that is part of the intervention being evaluated.

\subsection{Annotation layer: skill taxonomy and step tagger}
\label{sec:taxonomy}
To enable process-level evaluation of delegation decisions ---
specifically, whether a delegation was sent to a peer well-suited to
the sub-task --- DecisionBench ships an annotation layer. The
annotation layer is part of the benchmark's analysis machinery, not
a proposed orchestration method; methods are free to use it, ignore
it, or replace it with their own.

A fixed 7-skill taxonomy
(App.~\ref{app:taxonomy}, Table~\ref{tab:skills}) is frozen prior to
any tagging: tool-call schema adherence (BFCL$_{\text{primary}}$);
multi-turn state tracking ($\tau$); domain-policy compliance ($\tau$);
information retrieval (GAIA, $\tau$); multi-step reasoning (GAIA);
numerical computation (GAIA, $\tau$); long-input handling (GAIA).
Each skill has at least one primary suite, so the taxonomy exercises
observable behavior on every cell. A deterministic rule-based step
tagger (App.~\ref{app:tagging}, version-pinned) assigns at most one
skill to each step of an agent trajectory using trace-only signal
(tool name, finish reason, input-token threshold $\geq 15$K,
refusal-phrase regex), with no LLM judgment. The tagger is the basis
for two evaluation primitives in the metric suite: per-(model,
skill) Stage-1 pass-rate statistics (which define the
``Stage-1-best peer for a skill'' used in fidelity and ceiling
metrics) and the dominant-skill assignment for each Stage-2
delegation context. An emergent-taxonomy audit
(App.~\ref{app:emergent-taxonomy}) finds that $94.5\%$ of a
free-form LLM-judge re-tagging on a sub-sample maps cleanly to one
of the seven labels.

\subsection{Metrics}
\label{sec:metrics}
DecisionBench reports a multi-axis profile per orchestration method,
spanning outcome and process:

\textbf{Quality / cost / latency.} Per-suite native quality (GAIA
exact match, $\tau$-bench pass\textsuperscript{$k$}, BFCL AST match),
cost in USD per task at freeze-date OpenRouter pricing, and
wall-clock latency.

\textbf{Quality-cost trade-off.} 2D Pareto hypervolume in
$(q, -\text{cost})$ space against a fixed cost reference
($1.05\!\times$ max observed mean cost per benchmark), with
paired-bootstrap 95\% CIs ($n_{\text{boot}}{=}5000$) on the
difference vs.\ a designated baseline condition.

\textbf{Delegation rate.} Mean \texttt{call\_model} invocations per
task, per (model, suite, condition).

\textbf{Delegation fidelity@$k$.} For every observed
\texttt{call\_model} invocation, the rule-based tagger infers the
dominant skill of the orchestrator's pre-call trajectory; we rank
candidate peers by their Stage-1 pass rate on that skill and report
the share of delegations that pick one of the top-$k$ ($k\in\{1,3\}$).
This is a process-level routing-quality signal that does not depend
on the final task outcome.

\textbf{Vendor self-preference.} For each delegation we record the
(orchestrator vendor, peer vendor) pair and compare the observed
same-vendor share against the chance-level rate implied by pool
composition (a vendor with $k$ peers in an $N$-pool has expected
same-vendor share $(k{-}1)/(N{-}1)$).

\textbf{Counterfactual-delegation ceiling.} For each blind task we
compute the most-optimistic outcome if the orchestrator had
delegated the entire task to its Stage-1-best peer on the inferred
dominant skill, with the peer answering at its empirical Stage-1
pass rate. This bound is an explicit upper envelope under stated
assumptions (App.~\ref{app:ceiling-assumptions}) and serves as
unrealized-headroom information for future methods.

Paired-bootstrap CIs throughout are matched on task id, with
$\tau$-bench task ids prefixed by env shard so airline-task-0 is
distinct from retail-task-0. The full per-task records
($\sim$23k rows under the reference sweep in \S\ref{sec:setup}) are
released as \texttt{records.jsonl.gz} so every numerical claim is
reproducible from a single analysis script.

\section{Reference Awareness Conditions}
\label{sec:conditions}

The following five conditions are not part of the benchmark
definition; they are reference interventions instantiated on
DecisionBench to demonstrate how peer-awareness mechanisms are
evaluated and to seed comparisons for future methods. Each
intervention is fully specified by (i) what \texttt{read\_profile}
returns (the description content), and (ii) whether the same
content is also preloaded into the orchestrator's system prompt
(the delivery channel).

\paragraph{A concrete profile-card instance.}
\label{sec:concrete-card}
To ground the construction details that follow, we first present a
representative artifact: the abridged \texttt{C2} card for
\texttt{claude-opus-4.7}, rendered exactly as it would appear in the
orchestrator's context window.
\begin{quote}\small\vspace{-3pt}
\texttt{---}\\
\texttt{model: claude-opus-4.7;\ variant: c2\_static;\ tagger: v2.0-2026-05-01;}\\
\texttt{n\_tasks: 105;\ benchmarks: [bfcl, gaia, tau-bench]}\\
\texttt{---}\\
\textbf{claude-opus-4.7 --- derived skill profile (C2).}
Generated automatically from 105 Stage-1 tasks across 3 benchmarks.
No LLM judgment; reproducible from trace zips alone.\\[3pt]
\textbf{Strengths.} Domain-policy compliance (rank 2/10, $9/11\,{=}\,82\%$ pass,
\$0.069/success); numerical computation (3/11, $27/34\,{=}\,79\%$,
\$0.163/success); tool-call schema adherence (5/11, $41/52\,{=}\,79\%$).\\[1pt]
\textbf{Weaknesses.} Long-input handling (rank 2/8, $12/22\,{=}\,55\%$,
\$1.96/success); information retrieval (4/11, $35/46\,{=}\,76\%$);
multi-turn state tracking (5/11, $57/73\,{=}\,78\%$).\\[1pt]
\textbf{All measured skills.} {\footnotesize\setlength{\tabcolsep}{3pt}\begin{tabular}{lrrrrr}\toprule Skill & pass & $n$ & avg steps & avg out-tok & \$/task \\\midrule Domain-policy & $82\%$ & 11 & 1.2 & 256 & 0.056 \\Information retrieval & $76\%$ & 46 & 3.7 & 453 & 0.142 \\Multi-turn state tracking & $78\%$ & 73 & 8.3 & 504 & 0.258 \\\ldots & & & & & \\ \bottomrule\end{tabular}}\vspace{-3pt}
\end{quote}
The \texttt{C1} variant replaces the strength/weakness sentences
with curated vendor-claim citations and curator notes; \texttt{C3}
replaces them with two judge-written prose paragraphs (Grok-4,
Llama-4-Maverick) preserved side-by-side. All three variants share
the $\langle$YAML frontmatter, strengths, weaknesses, all-skills
table$\rangle$ skeleton; the full set of $11{\times}3{=}33$ cards is
released under \texttt{profile\_cards/}.

\subsection{Blind delegation baseline}
\label{sec:blind}
\texttt{blind} exposes only \texttt{call\_model} with the peer-name
list and no skill information; \texttt{read\_profile} is not
attached. The orchestrator delegates entirely from priors over peer
names. This is the no-information reference against which the four
awareness interventions below are compared. Empirically (see
Table~\ref{tab:headline}) blind delegation rates are 0.02 on
$\tau$-bench, 0.17 on BFCL, and 0.41 on GAIA per task, so blind
quality is within $\sim$1--2~pp of solo on $\tau$-bench and BFCL and
serves as the operative no-delegation baseline on those two suites;
a full Stage-2 solo (no-\texttt{call\_model}-tool) cell is queued
for a future release.

\subsection{Profile-card content variants}
\label{sec:profile-flavors}
The three content variants share schema and differ only in
construction; all are rendered as markdown for an agent's context
window. Per-(model, skill) Stage-1 static metrics (pass rate, mean
tokens, mean steps, mean latency, cost-per-success) are the
arithmetic primitive on which \texttt{C2} is built, and which
\texttt{C3} judges read alongside raw traces.

\texttt{C1} (structured-rubric anchor): a curated per-skill profile
assembled from public sources only (vendor model cards, public
leaderboards~\citep{swebenchproboard,bfcl,skillsbench,lmarena}) with
provenance markers; single-curator, pre-registered at a dated git
hash, treated as a structured-rubric anchor rather than a validated
human baseline.

\texttt{C2} (static-metric): deterministic statistics over the
agent's own Stage-1 traces (per-(model, skill) pass rate, tokens,
steps, latency, cost-per-success, percentile rank), end-to-end
reproducible from released zips with tagger version pinned. No LLM
judgment, so reviewer disagreement collapses to verifiable
arithmetic.

\texttt{C3} (dual-LLM-judge): free-form skill summaries from two
out-of-pool judges (xAI Grok-4, Meta Llama-4-Maverick) that read
each candidate's Stage-1 traces independently; both judgments
preserved side-by-side, not synthesized. Inter-judge
Spearman~$\bar\rho$ (App.~\ref{app:c3-judges}) is itself a
measurable sub-metric.

\subsection{Delivery channels: preloaded versus on-demand}
\label{sec:delivery}
The same description content can reach an orchestrator through two
qualitatively different channels: pasted into the system prompt
before the task starts (preloaded), or retrievable through a
\texttt{read\_profile} tool call mid-task (on-demand). The aware
conditions span both axes:
\begin{itemize}\setlength\itemsep{0pt}
\item \texttt{aware-c1}, \texttt{aware-c2}, \texttt{aware-c3} ---
 \texttt{call\_model} + \texttt{read\_profile} serving the matching
 \texttt{C1}/\texttt{C2}/\texttt{C3} card on demand, with the same
 description \emph{also} preloaded in the system prompt.
\item \texttt{aware-tool-only} --- same tools as \texttt{aware-c2}
 (the \texttt{C2} card is what \texttt{read\_profile} returns), but
 with the preloaded peer description suppressed. This is a
 one-variable ablation against \texttt{aware-c2}, pinning content
 at \texttt{C2} and isolating the delivery-channel axis.
\end{itemize}
Cards are served on demand so which cards the agent reads is
itself measurable behavior.

\subsection{Stage-2 condition summary}
\label{sec:stage2}
Table~\ref{tab:conditions} summarizes the five conditions on the
content$\times$delivery axes.

\begin{table}[h]
\centering
\small
\setlength{\tabcolsep}{6pt}
\begin{tabular}{lllc}
\toprule
Condition & \texttt{call\_model} & \texttt{read\_profile} returns & preloaded \\
\midrule
\texttt{blind}      & yes & (tool not attached) & no \\
\texttt{aware-c1}     & yes & \texttt{C1} card    & \texttt{C1} description \\
\texttt{aware-c2}     & yes & \texttt{C2} card    & \texttt{C2} description \\
\texttt{aware-c3}     & yes & \texttt{C3} card    & \texttt{C3} description \\
\texttt{aware-tool-only} & yes & \texttt{C2} card    & no \\
\bottomrule
\end{tabular}
\caption{Five reference conditions instantiated on DecisionBench,
along the content axis (\texttt{C1} curated rubric / \texttt{C2}
deterministic statistics / \texttt{C3} dual LLM-judge) and the
delivery axis (preloaded into the system prompt versus retrieved
on demand through \texttt{read\_profile}). \texttt{aware-tool-only}
is the one-variable ablation against \texttt{aware-c2}.}
\label{tab:conditions}
\end{table}

\section{Experimental Setup}
\label{sec:setup}

We characterize the substrate by running the five reference
conditions on the eleven-model pool, yielding
$11 \text{ models} \times 3 \text{ benchmarks} \times 5 \text{ conditions} = 165$
cells (220 run zips after $\tau$-bench airline/retail sharding). Each
cell evaluates 133 GAIA tasks, 132 $\tau$-bench tasks, or 160 BFCL
tasks against the substrate metric suite (\S\ref{sec:metrics}); total
$n{=}23{,}375$ task instances. All scoring is post-hoc against the
canonical metrics. Compute budget and per-model spend are detailed
in App.~\ref{app:compute}.

For statistical inference we use two estimators. Paired-bootstrap
95\% CIs on $\Delta q$ vs.\ \texttt{blind} are matched on task id
($n_{\text{boot}}{=}5000$, basic / Hall-reflected). For the global
test that any aware condition shifts mean quality, we fit a
random-intercept mixed-effects model $q \sim \text{cond} +
(1|\text{agent}\times\text{benchmark})$ on the $23{,}375$ task rows
with \texttt{statsmodels}~\citep{statsmodels}; the full output is in
App.~\ref{app:mixedlm}. Random slopes for \texttt{cond} did not
converge under any of four optimizers we tried, so the fit is
random-intercept only; this is conservative since unmodeled
task-level variance inflates residual SEs.

\paragraph{Released artifacts.}
We release: the DecisionBench substrate
(\texttt{decision\_bench/} task / pool / interface / metric code);
the annotation layer (\texttt{tagger.py} v2.0, frozen 7-skill
taxonomy, $200$-step emergent-audit CSV); the reference intervention
suite (\texttt{profile\_cards/} with all 33 cards in three variants,
plus the \texttt{aware-tool-only} ablation harness); the analysis
pipeline; and 220 per-condition run archives spanning
$\sim$23{,}000 task-level traces. The C2 cards are reproducible
from the released zips via \texttt{tools/build\_c2\_profile.py};
the C3 cards via \texttt{tools/build\_c3\_profile.py} ($\sim$\$10--\$20
of judge inference).

\section{Results}
\label{sec:results}

We organize results around six questions DecisionBench is designed
to answer about a peer-awareness intervention:
(\S\ref{sec:results-quality}) does it improve mean end-task quality?
(\S\ref{sec:results-heterogeneity}) where is the signal hidden in
the aggregate, across agents and suites?
(\S\ref{sec:results-decomposition}) does delivery channel or
description content matter more?
(\S\ref{sec:results-fidelity}) does it improve which peer the
orchestrator chooses?
(\S\ref{sec:results-vendor}) what failure modes does the benchmark
surface?
(\S\ref{sec:results-ceiling}) how much unrealized headroom remains
for future methods?

\subsection{End-task quality is mostly flat across awareness conditions}
\label{sec:results-quality}\label{sec:results-headline}

\begin{figure}[t]
\centering
\includegraphics[width=\textwidth]{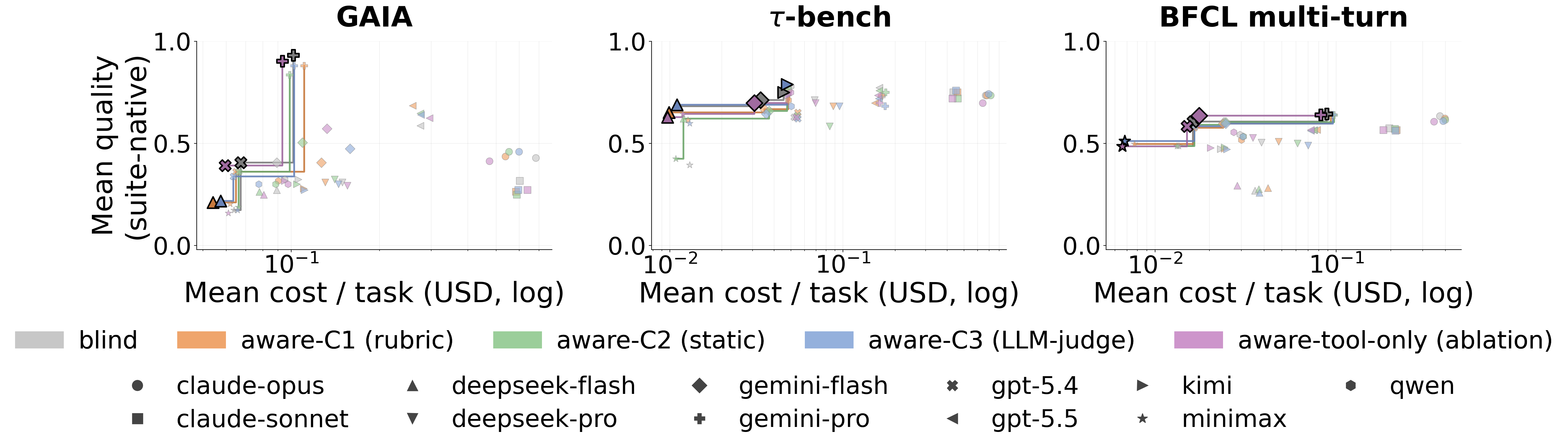}
\caption{Quality--cost Pareto frontier per benchmark
(GAIA: general tool-use; $\tau$-bench: tool--agent--user dialogue;
BFCL multi-turn: function-calling), 5 conditions overlaid.
One marker per agent, one color per condition; step curves trace each
condition's frontier across the 11 agents. Aggregate frontiers are
largely superimposed; per-agent Pareto-favorable movement
(Fig.~\ref{fig:pareto-wins}) is where awareness effects concentrate. Aware-tool-only (purple) tracks the
blind frontier on GAIA / BFCL and runs at lower mean cost than
the preloaded-aware variants because it avoids the per-call $\sim$30-line
peer-description augmentation in the system prompt.}
\label{fig:pareto}
\end{figure}

Mean end-task quality is statistically indistinguishable across the
four awareness conditions, on every suite. Routing fidelity, in
contrast, produces a clear ordering across the same five conditions
(\S\ref{sec:results-fidelity}); the dissociation is the
methodological payoff of measuring process- and outcome-level
metrics together. Table~\ref{tab:headline} reports per-condition
aggregates; Fig.~\ref{fig:pareto} shows the underlying quality--cost
frontiers.

Awareness improves routing precision. Awareness lifts routing fidelity@1 in three
of four conditions: blind 14.2\% $\to$ aware-c1 7.5\%, aware-c2
20.8\%, aware-c3 15.5\%, aware-tool-only 29.5\%. On-demand tool
access drives the largest gain; partial-preload C2 and C3 capture
roughly half; only the rubric variant (C1) underperforms blind.
Mean quality is flat across all four awareness conditions: $\Delta$q
is $\{-1.2,-0.4,-1.4,+0.1\}$~pp on GAIA,
$\{+0.1,-2.3,-0.3,-0.1\}$~pp on $\tau$-bench, and
$\{-0.5,-0.4,-0.8,+0.3\}$~pp on BFCL for aware-c1/c2/c3/tool-only. A mixed-effects
regression $q \sim \text{cond} + (1\,\vert\,\text{agent}\!\times\!\text{benchmark})$
on 23{,}375 task rows confirms all four aware coefficients are
indistinguishable from blind
($\beta \in \{-0.005,-0.010,-0.008,+0.001\}$, all $p \geq 0.21$);
the largest pairwise contrast is C2 vs.\ tool-only
$\Delta\hat\beta{=}-0.011$, $p{=}0.18$
(full table App.~\ref{app:hypervolume}). This dissociation ---
fidelity moves cleanly across conditions while mean quality does
not --- is itself a methodological finding: process-level metrics
detect orchestration design differences that quality-only evaluation
misses. Whether the routing-fidelity gain propagates to outcome at
larger sample sizes or under multi-step delegation is open work
(\S\ref{sec:limitations}).

\paragraph{Negative $\Delta$HV reflects a cost-band shift, not
quality regression.}
Six of nine $\Delta$HV CIs (paired bootstrap, basic; full table in
App.~\ref{app:hypervolume}) exclude zero, all in the negative
direction --- four GAIA aware variants plus two $\tau$-bench cells
(aware-c2 $-0.011$ $[-.019,-.005]$; aware-tool-only $-0.017$
$[-.027,-.014]$). These are driven entirely by the per-call
$\sim$30-line peer-description augmentation in the system prompt
translating the same quality cluster to a higher cost band
($+0.02$--$0.05$/task), consistent with the near-flat $\Delta$q
coefficients from the mixed-effects fit above. \texttt{aware-tool-only},
which avoids this prompt augmentation, is bootstrap-flat or weakly
positive on every suite. BFCL is bootstrap-flat across all conditions.

\paragraph{Reading the frontiers (Fig.~\ref{fig:pareto}).}
Per-suite reading: GAIA preloaded variants translate the same quality
cluster to a higher cost band ($+0.02$--$0.05$/task); on $\tau$-bench,
aware-c2 and aware-tool-only drift lower-tier agents downward (the
two cells with HV CIs excluding zero); BFCL is bootstrap-flat.
Aggregate frontiers are largely superimposed.

\begin{table}[h]
\centering
\small
\setlength{\tabcolsep}{4pt}
\begin{tabular}{llrrrr}
\toprule
Bench & Cond.\ & mean q & $\Delta$q vs.\ blind & mean \$ & dlg/task \\
\midrule
\multirow{5}{*}{GAIA}
 & blind      & 0.407 & ---   & 0.211 & 0.41 \\
 & aware-c1     & 0.395 & $-$0.012 & 0.194 & 0.28 \\
 & aware-c2     & 0.403 & $-$0.004 & 0.197 & 0.31 \\
 & aware-c3     & 0.393 & $-$0.014 & 0.205 & 0.29 \\
 \rowcolor{rowhighlight} & \textbf{aware-tool-only} & 0.408 & $+$0.001 & 0.199 & 0.19 \\
\midrule
\multirow{5}{*}{$\tau$-bench}
 & blind      & 0.695 & ---   & 0.155 & 0.02 \\
 & aware-c1     & 0.696 & $+$0.001 & 0.159 & 0.01 \\
 & aware-c2     & 0.672 & $-$0.023 & 0.164 & 0.02 \\
 & aware-c3     & 0.692 & $-$0.003 & 0.162 & 0.01 \\
 \rowcolor{rowhighlight} & \textbf{aware-tool-only} & 0.694 & $-$0.001 & 0.152 & 0.00 \\
\midrule
\multirow{5}{*}{BFCL}
 & blind      & 0.536 & ---   & 0.083 & 0.17 \\
 & aware-c1     & 0.531 & $-$0.005 & 0.089 & 0.12 \\
 & aware-c2     & 0.533 & $-$0.004 & 0.090 & 0.13 \\
 & aware-c3     & 0.529 & $-$0.008 & 0.089 & 0.12 \\
 \rowcolor{rowhighlight} & \textbf{aware-tool-only} & 0.539 & $+$0.003 & 0.076 & 0.06 \\
\bottomrule
\end{tabular}
\caption{Stage-2 aggregates: mean per-task quality, $\Delta$q vs.\
blind, mean cost (USD), and mean delegation count per task, averaged
across the 11 agents. The aware-tool-only condition matches or beats
blind on quality at lower mean cost and lower delegation count, the
most parsimonious of the five conditions on every dimension. Per-suite
delegation-rate context appears in \S\ref{sec:tau-bfcl-deleg}.}
\label{tab:headline}
\end{table}

\paragraph{Where delegation actually fires.}
\label{sec:tau-bfcl-deleg}
The three substrates differ sharply in how often \texttt{call\_model}
fires, which qualifies how to read Table~\ref{tab:headline}. GAIA
invites delegation (15--30\% of tasks; blind $0.42$/task,
\texttt{aware-tool-only} $0.19$); BFCL sits in the middle
($\sim$10\%); on $\tau$-bench it is near zero across all conditions
(89 invocations in $5{\times}1452$ task-slots), as agents adhere to
the per-domain Agent Policy regardless of how peer information is
presented. The flat $\tau$-bench numbers in Table~\ref{tab:headline}
therefore reflect priming from the system prompt rather than
delegation behavior. Where delegation does fire, aware variants
delegate less than blind, which is part of why
\texttt{aware-tool-only} reaches quality parity at lower cost.

\begin{figure}[h]
\centering
\includegraphics[width=0.92\textwidth]{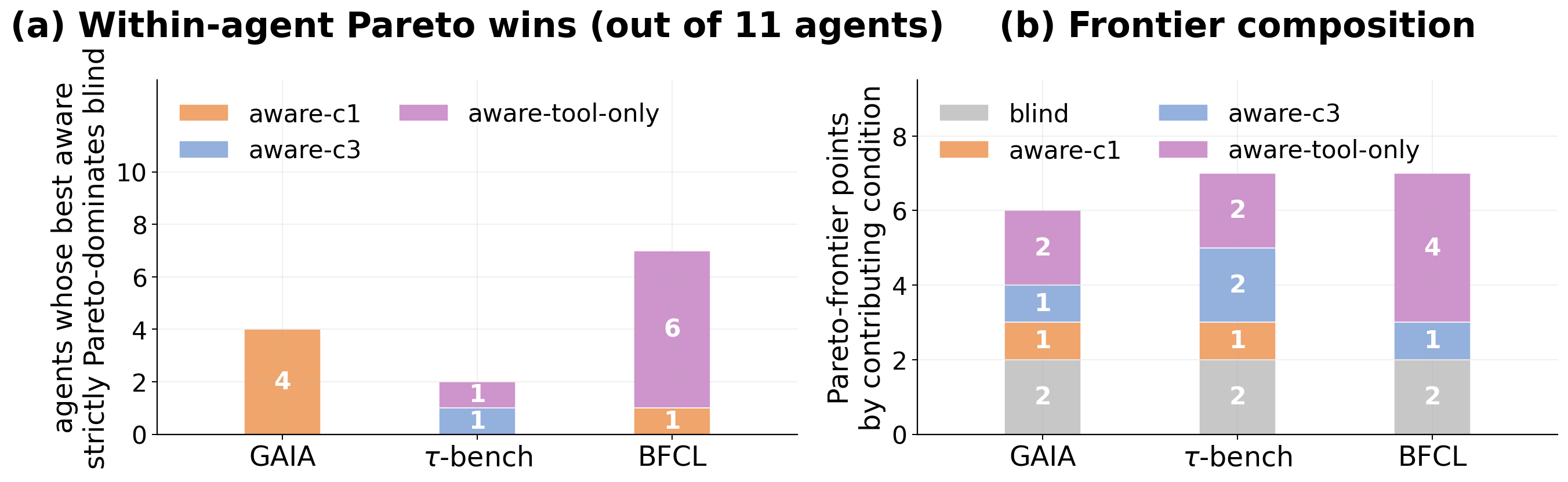}
\caption{(a) Agents (of 11) whose best aware variant strictly
Pareto-dominates blind, by aware condition. (b) Pareto-frontier
composition per benchmark; \texttt{aware-tool-only} contributes 4 of 7
BFCL frontier points.}
\label{fig:pareto-wins}
\end{figure}

\paragraph{Per-agent Pareto wins (Fig.~\ref{fig:pareto-wins}).}
Counting agents whose best aware variant strictly Pareto-dominates
the blind cell:
BFCL 7/11 (6 from \texttt{aware-tool-only}), GAIA 4/11
(all from \texttt{aware-c1}), $\tau$-bench 2/11. Frontier composition
agrees: on BFCL, 4 of 7 frontier points (57\%) are
\texttt{aware-tool-only}, the modal condition. Aggregate frontier
flatness coexists with within-agent Pareto-favorable movement on
BFCL (7/11 agents) and GAIA (4/11) --- the granularity at which the
orchestration recommendation actually applies.

\subsection{Per-agent and per-suite heterogeneity}
\label{sec:results-heterogeneity}\label{sec:results-per-agent}
The aggregate hides per-agent heterogeneity (App.~\ref{app:capability-orch}):
on GAIA, Gemini-3-Flash gains $+16.5$~pp (aware-tool-only) and
Gemini-3.1-Pro regresses $-9.8$~pp (aware-c2); a concave parabolic
fit on GAIA / BFCL identifies a mid-capability regime where awareness
helps most. In the per-variant breakdown
(App.~\ref{app:per-agent-bars}), \texttt{aware-tool-only} is at the
top of the bar group for 8 of 11 agents, motivating the
content-vs-delivery decomposition we develop next.

\paragraph{Cross-suite generalization is weak.}
\label{sec:results-cross-suite-early}
Across 11 agents (using each agent's best aware lift over blind per
suite): Spearman $\rho(\Delta_{\text{GAIA}}, \Delta_{\tau}) = -0.38$,
$\rho(\Delta_{\text{GAIA}}, \Delta_{\text{BFCL}}) = +0.48$,
$\rho(\Delta_{\tau}, \Delta_{\text{BFCL}}) = -0.21$. The negative
GAIA--$\tau$ correlation is interpretable: GAIA rewards decomposition;
$\tau$-bench rewards policy adherence (delegation pulls the agent
off-script). Orchestration ability is at least partly suite-specific.

\subsection{Delivery channel dominates description content}
\label{sec:results-decomposition}
\begin{figure}[h]
\centering
\begin{minipage}[c]{0.62\textwidth}
\centering
\includegraphics[width=\linewidth]{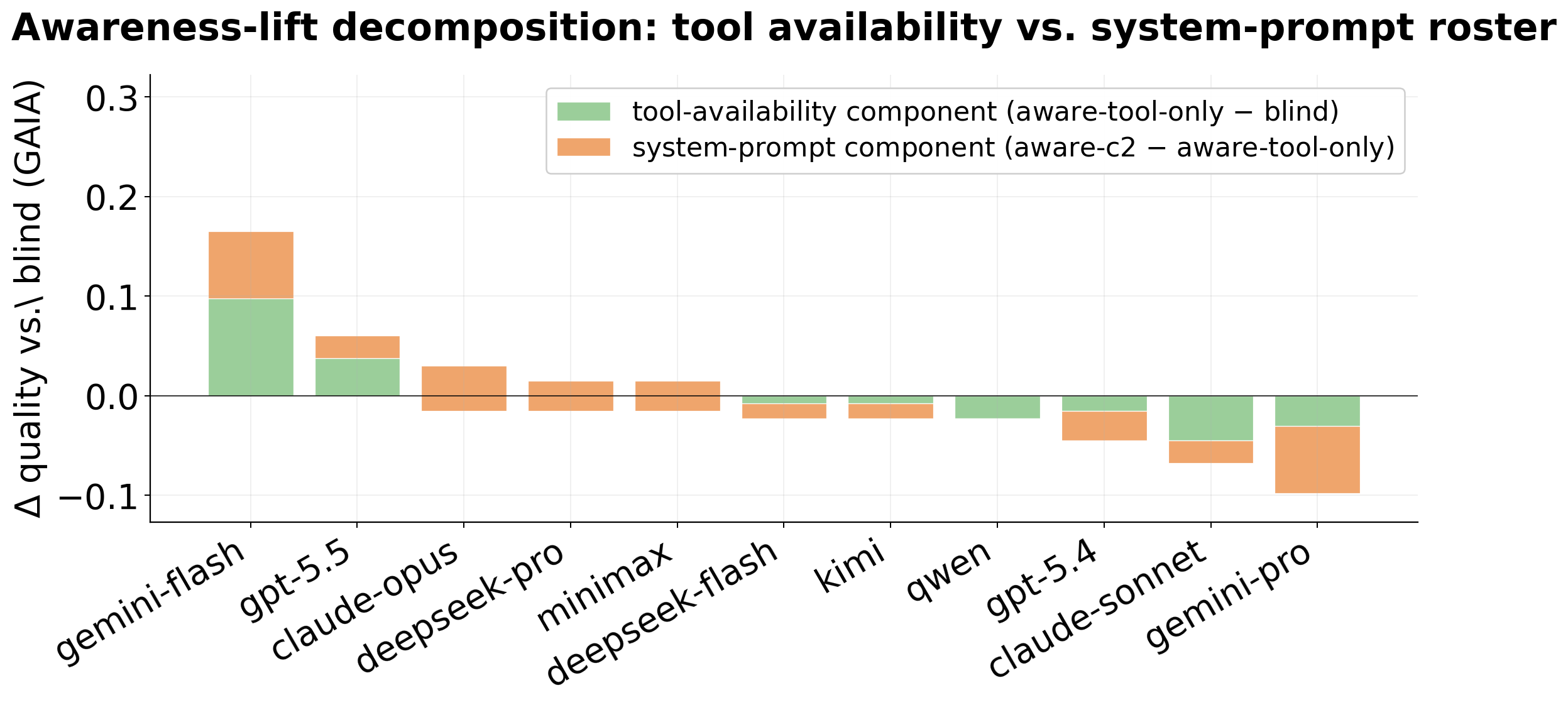}
\end{minipage}\hfill
\begin{minipage}[c]{0.36\textwidth}
\centering
\small
\setlength{\tabcolsep}{4pt}
\begin{tabular}{lrr}
\toprule
Condition & $n$ & fid@1 \\
\midrule
blind       & 339 & 14.2\% \\
aware-c1     & 227 & 7.5\% \\
aware-c2     & 250 & 20.8\% \\
aware-c3     & 233 & 15.5\% \\
aware-tool-only & 122 & 29.5\% \\
\bottomrule
\end{tabular}
\\[4pt]
{\footnotesize\itshape Tab.~2: Fidelity@1 is the share of delegations that pick the Stage-1-best peer.}
\end{minipage}
\caption{Left: per-agent decomposition of the aware-c2 GAIA lift
over blind into the tool-availability component (purple,
\texttt{aware-tool-only}~$-$~blind) and the system-prompt component
(green, aware-c2~$-$~\texttt{aware-tool-only}). System-prompt component
is negative on most agents; the few positive tool-availability
agents (notably Gemini-3-Flash) are reduced by the system-prompt
augmentation. Right: delegation fidelity per condition. Adding
on-demand tool access doubles fidelity (14.2\%~$\to$~29.5\%); the same
information re-rendered as a preloaded description recovers less than half of the gain
(C1 7.5\%, C2 20.8\%, C3 15.5\%).}
\label{fig:decomposition}
\end{figure}

The aware intervention bundles three components: the \texttt{call\_model} tool,
the \texttt{read\_profile} tool, and a peer description appended to the
system prompt. The aware-tool-only condition keeps the first two and
suppresses the third, allowing us to decompose the awareness improvement into a
tool-availability component ($q_{\text{aware-tool-only}} -
q_{\text{blind}}$) and a system-prompt component
($q_{\text{aware-c2}} - q_{\text{aware-tool-only}}$).
Figure~\ref{fig:decomposition} shows the per-agent stacked decomposition on
GAIA; the bars are paired-bootstrap significant where indicated.

Across the 33 (agent, benchmark) cells with paired data, the
system-prompt component is significantly $> 0$ on zero cells and
significantly $< 0$ on 4~cells; the tool-availability component is $>
0$ on 4~cells and $< 0$ on 2~cells. So the only quality-side signal
in the decomposition is on the tool-availability axis, and even that
signal is small in absolute pp. The asymmetry --- tool-availability
moves quality on a few cells while system-prompt augmentation does
not move it positively on any --- is the quality-side echo of the
much larger fidelity gap reported in \S\ref{sec:results-fidelity}.

\subsection{Awareness improves delegation fidelity}
\label{sec:results-fidelity}

For each \texttt{call\_model} invocation in the data we infer the dominant
skill the orchestrator was working on (the modal skill-tag in the
trace up to that step), look up each candidate peer's Stage-1 pass rate on
that skill from the C2 cards, and rank them. Fidelity@1 is the share of
delegations that pick the top-ranked peer for the relevant skill;
fidelity@3 is the share that picks one of the top three.
The per-condition table sits inside Fig.~\ref{fig:decomposition} (right);
fidelity@3 numbers and per-cell breakdowns are released as
\texttt{analysis/fidelity\_per\_cond.csv} (App.~\ref{app:fidelity-per-cell}).

The data is most consistent with preloaded peer descriptions being a
less effective delivery channel than on-demand tool access, and with
the C1 rubric variant --- prose claims with vendor-name provenance ---
providing weaker delegation cues than the trace-grounded C2/C3
variants. The ordering across conditions makes this concrete:
on-demand tool access (blind $\to$ aware-tool-only) doubles fidelity
$14.2\%~\to~29.5\%$; preloading the same information lands at
$7.5\%~/~20.8\%~/~15.5\%$ (C1~/~C2~/~C3), with only C1 strictly
underperforming blind. Preloaded variants capture between roughly
$9\%$ (C3) and $43\%$ (C2) of the on-demand fidelity gain over
blind; C1 captures none of it. The gap likely reflects two effects: agents reading cards on demand
exercise discrimination that a preloaded list short-circuits, and
preloaded lists surface peer names as pattern-matching cues that
substitute for reading the cards themselves. Methodological
recommendation: surface peer-skill information through tools, not
preloaded prompt sections.

\subsection{Cross-vendor self-preference}
\label{sec:results-vendor}\label{sec:analyses}

\begin{figure}[h]
\centering
\includegraphics[width=\textwidth]{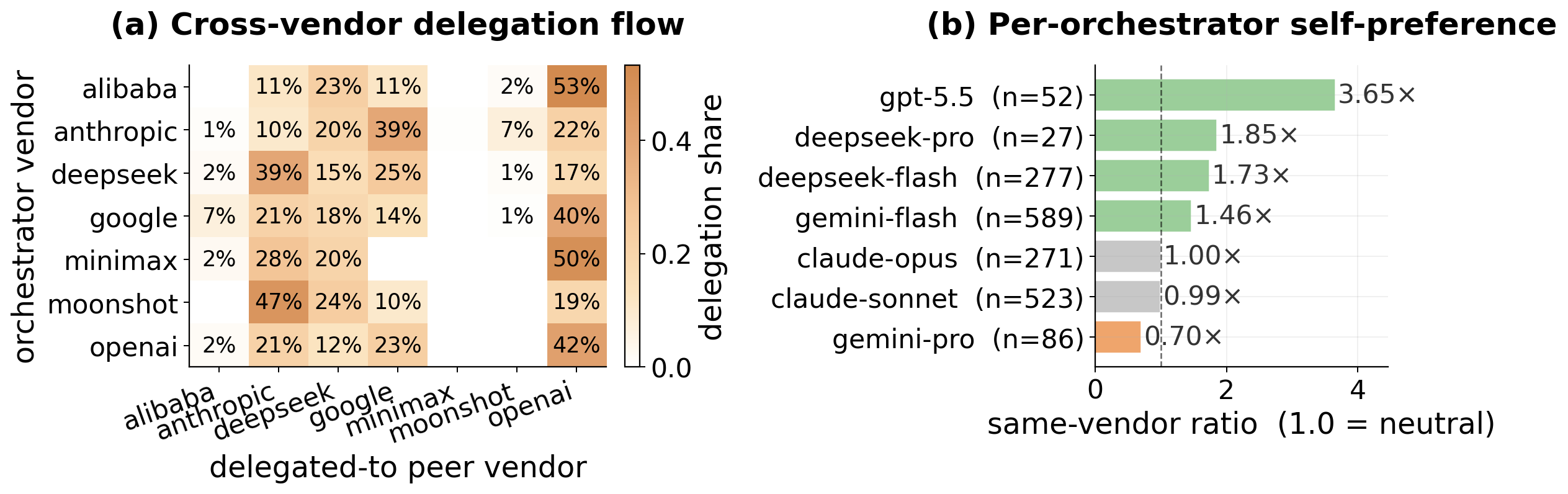}
\caption{(a) Cross-vendor delegation flow (aware-* aggregated):
rows are orchestrator vendors, columns are peer vendors, cells are
row-normalized percentages. (b) Per-orchestrator self-preference
ratio (actual same-vendor share divided by chance). Bars right of
$1.0\times$ over-prefer same-vendor (green); left of $1.0\times$
under-prefer (orange).}
\label{fig:vendor-heatmap}
\end{figure}

For each delegation we record (orchestrator vendor, peer vendor) and
compare the share of same-vendor delegations against the chance-level rate
implied by the pool composition (a vendor with $k$ models in the
$N{=}11$~pool has expected same-vendor share $(k{-}1)/(N{-}1)$).
Several agents over-delegate at $1.5{-}3.7\times$ chance: Gemini-3-Flash
$1.48\times$ ($n{=}594$), DeepSeek-V4-Flash $1.53\times$ ($n{=}594$),
DeepSeek-V4-Pro $1.85\times$ ($n{=}27$), GPT-5.5 $3.65\times$ ($n{=}52$).
Anthropic is neutral; Gemini-3.1-Pro anti-prefers same-vendor at
$0.70\times$. The full vendor-by-vendor flow (Fig.~\ref{fig:vendor-heatmap})
shows Anthropic and Google peers absorbing a disproportionate share of
cross-vendor delegations from every other vendor. This is the
delegation-tool analogue of LLM-as-judge self-preference
bias~\citep{zheng2023judging}.

\subsection{Counterfactual-delegation ceiling: room for future methods}
\label{sec:results-ceiling}\label{sec:results-cross-suite}

To bound the room for improvement: for each blind task, if the agent
had delegated to the Stage-1-best peer on the inferred dominant skill
(peer answers at its Stage-1 pass rate), what's the most-optimistic
outcome? Aggregating across the 11 agents:

\begin{center}
\small
\setlength{\tabcolsep}{6pt}
\begin{tabular}{lrrr}
\toprule
Benchmark & blind actual & counterfactual ceiling & gap \\
\midrule
GAIA     & 0.407 & 0.675 & $+0.269$ \\
$\tau$-bench & 0.695 & 0.848 & $+0.153$ \\
BFCL     & 0.536 & 0.849 & $+0.313$ \\
\bottomrule
\end{tabular}
\end{center}

\begin{figure}[h]
\centering
\includegraphics[width=0.97\textwidth]{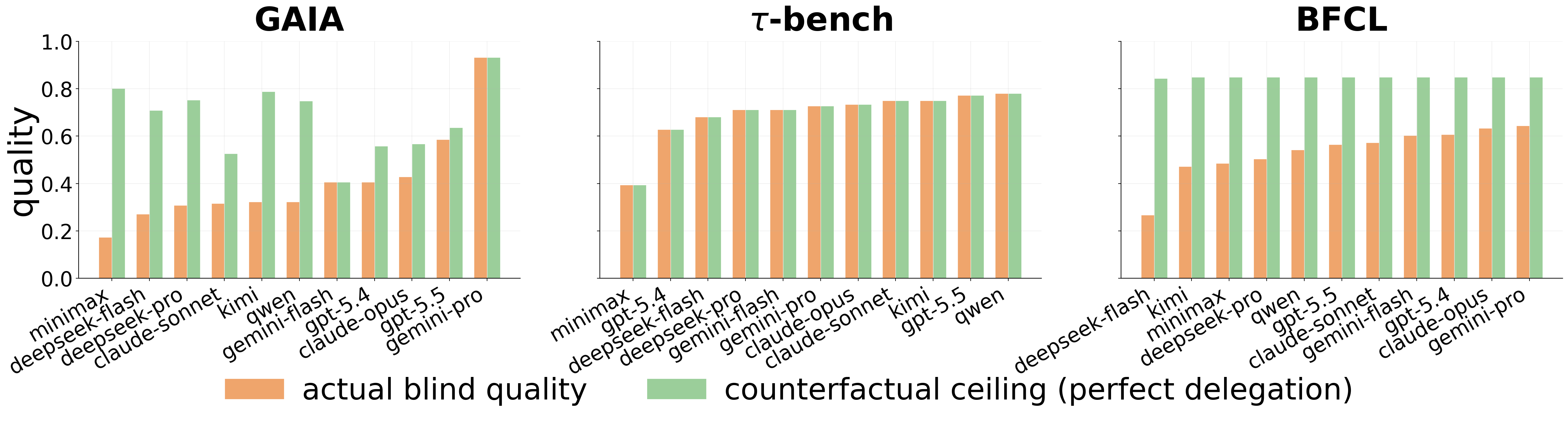}
\caption{Counterfactual-delegation ceiling per (model, benchmark): actual
blind quality (purple) vs.\ the upper bound from delegating every task to
the Stage-1-best peer for the inferred dominant skill (green). The gap is
largest for weak / mid-tier agents (MiniMax-M2.5 on GAIA: $0.17{\to}0.80$,
$+63$~pp possible) and near zero for already-saturated cells (Gemini-3.1-Pro
on GAIA at $0.93$). Current awareness deliveries capture only a
fraction of this room on quality, leaving substantial room for
future interventions to close.}
\label{fig:ceiling}
\end{figure}

The ceiling sits 15--31~pp above observed performance on every
benchmark; current awareness deliveries capture only a fraction of
this room on quality, with most unrealized headroom on weak /
mid-capability agents (per-agent breakdown in Fig.~\ref{fig:ceiling}).

\paragraph{Case studies.}
\label{sec:case-studies}
\textit{Gemini-3-Flash on GAIA: the mechanism in action.}
$q_{\text{blind}}{=}0.406 \to 0.571$ ($+16.5$~pp) under
\texttt{aware-tool-only}, the largest single-cell improvement in our
experiments. The same information delivered as a preloaded
description (\texttt{aware-c2}) gives only $+9.8$~pp; \texttt{aware-c1}
gives $0$. Fidelity@1 tracks the ordering (blind~$11\%$, c1~$4\%$,
c2~$14\%$, tool-only~$33\%$): the agent makes better peer choices
on-demand. \textit{Claude-Sonnet-4.6 on GAIA: a clean negative case ---
a credibility check that the benchmark can produce a negative result
when the mechanism does not engage.} Blind $q{=}0.316$; all four
aware variants regress, including on-demand
($\{-5.3,-6.8,-4.5,-4.5\}$~pp for c1/c2/c3/tool-only): moderate
blind quality does not guarantee a positive response if the
orchestrator's discrimination over peer cards is itself poor.
Three further cases --- MiniMax-M2.5's open ceiling that does not
translate; GPT-5.5's pool-leading $3.65\times$ same-vendor delegation
rate; MiniMax-M2.5 on $\tau$-bench winning by delegation suppression
rather than by routing precision --- appear in the released artifact.

\paragraph{Skill-level breakdown.}
\label{sec:results-skill}\label{sec:results-recovery}
By dominant skill, aware variants improve where peer diversity
matters most: multi-step reasoning ($+5.9$~pp tool-only, $n{=}85$)
and information retrieval ($+8/{+}11$~pp c1/c3); tool-call schema
adherence ($n{=}1384$) and long-input handling are essentially flat
--- skills where the orchestrator's own capability dominates and peer
information adds little. Per-skill detail and C3 inter-judge
agreement ($\bar\rho{=}0.54$, range $[0.31, 0.78]$):
App.~\ref{app:skill-lift-fig}, App.~\ref{app:c3-judges}.

\section{Discussion}
\label{sec:discussion}

\subsection{What DecisionBench measures beyond accuracy}
The benchmark's main payoff over accuracy-only evaluation is the
process-level signal it instruments. Across the reference sweep,
end-task quality is flat across the four awareness conditions
($|\beta|{\leq}0.010$, $p{\geq}0.21$), yet delegation
fidelity-at-$1$ ranges from $7.5\%$ (\texttt{aware-c1}) to $29.5\%$
(\texttt{aware-tool-only}) and vendor self-preference reaches
$3.65\times$ chance on one orchestrator
(\S\ref{sec:results-vendor}). A benchmark that only measured
quality would see no difference between any of the conditions; a
benchmark that did not record per-delegation (skill, peer, vendor)
metadata could not distinguish which peer was chosen, only how the
task ended. The fidelity / vendor-bias / ceiling axes also matter
for methods that explicitly target peer-selection quality
(learned routers, multi-step planners): an evaluation that scores
them only on mean accuracy would miss exactly the dimension they
are designed to move.

\subsection{Why better routing has not (yet) translated to quality}
The fidelity gain from on-demand tool access is real and large
(doubling over blind), and yet mean quality is flat. Several
mechanisms are consistent with the data. Delegation rates are
modest on every suite (0.02--0.41 per task) and near-zero on
$\tau$-bench, so even perfect peer-selection on the delegated calls
moves a small share of total task outcome.
The C3 judges' inter-rater Spearman~$\bar\rho{=}0.54$ suggests
peer descriptions themselves carry only partial signal about which
peer to pick on a given sub-task. And the per-agent decomposition
(\S\ref{sec:results-decomposition}) suggests preloaded descriptions
crowd attention away from the on-demand channel, partially
offsetting the gain. Closing the fidelity-quality gap is plausibly
about delegating more (when warranted), delegating multi-step, or
giving the orchestrator richer mid-task feedback --- all of which
DecisionBench can evaluate without further substrate changes.

\subsection{Future orchestration methods enabled by the benchmark}
Because the substrate (\S\ref{sec:design}) fixes only the tasks,
pool, interface, and metrics, methods that swap in different
peer-information generators, delivery channels, or delegation
policies can be evaluated head-to-head: learned routers
(\citep{routellm,routerbench}) plugged in as a different
\texttt{call\_model} policy, multi-step delegation with
per-sub-task peer choice, adaptive profile construction that
re-summarizes peers during the run, richer peer memories
accumulated across tasks, or heterogeneous pools that include
sub-30B and specialist models. The released annotation layer (frozen
7-skill taxonomy plus deterministic tagger) and process metrics
(fidelity@$k$, self-preference, counterfactual ceiling) carry over
unchanged, so cross-method comparisons remain anchored even as the
intervention space expands.

\section{Limitations}
\label{sec:limitations}

\textbf{Single-seed Stage-2.} Each Stage-2 cell is run once; we do
not characterize seed-to-seed variability beyond paired-bootstrap
CIs over tasks. \textbf{Single-curator C1.} The \texttt{C1} cards
are pre-registered at a dated git hash but reflect one curator's
synthesis; reviewer disagreement on \texttt{C1} cannot be checked
the way it can on \texttt{C2} (deterministic arithmetic) or
\texttt{C3} (two judges side-by-side). \textbf{$\tau$-bench
delegation is near-zero.} Agents adhere to domain policy regardless
of peer information; the $\tau$-bench numbers therefore reflect
system-prompt priming more than emergent delegation behavior, and
should not be used to compare orchestration methods in isolation.
\textbf{C3 judges saw Stage-1 outcomes.} The judges scored peer
traces with pass/fail information visible; an ablation that hides
outcomes from the judges would strengthen the C3 vs.\ C2
comparison. \textbf{Mixed-effects fit uses agent$\times$benchmark
random intercepts only.} Random slopes for \texttt{cond} did not
converge under four optimizers; the random-intercept fit is
conservative because unmodeled task-level variance inflates
residual SEs. \textbf{Cross-suite Spearman on $n{=}11$ is
suggestive only}; the absolute correlations
(\S\ref{sec:results-ceiling}) carry wide CIs. \textbf{Pool freeze
date.} The 11-model pool is pinned to 2026-04-29 OpenRouter
availability and pricing; absolute numbers shift under different
freeze dates, though the cross-condition design is robust.
\textbf{Existing-task reuse.} DecisionBench draws on GAIA,
$\tau$-bench, and BFCL rather than introducing new tasks; benchmark
contamination in the underlying suites flows through to our
measurements.

\section{Conclusion}
\label{sec:conclusion}\label{sec:repro-ethics}

DecisionBench is a reusable substrate for measuring emergent
delegation in long-horizon agentic workflows: a task suite, peer
pool, delegation interface, annotation layer, and a multi-axis
metric set that together let benchmark users compare orchestration
methods on routing fidelity, vendor self-preference, and
unrealized-headroom in addition to end-task quality. As an initial
characterization of the substrate, our reference intervention suite
shows that delivery channel dominates description content for
peer-awareness: on-demand tool access more than doubles routing
precision-at-$1$ over a no-information baseline at quality parity,
while preloaded peer descriptions capture less than half of that
gain. Mean end-task quality is flat across all four awareness
conditions, locating the orchestration signal in process metrics
that quality-only evaluation would miss, and a perfect-delegation
ceiling sits 15--31~pp above measured performance on every suite ---
substantial unrealized headroom for future methods. We release the
substrate, the annotation layer, the reference intervention suite
(33 cards in three variants plus the \texttt{aware-tool-only}
ablation), the analysis pipeline, and 220 per-condition run
archives so that learned routers, multi-step delegation, adaptive
profile construction, and richer peer pools can be evaluated
head-to-head on the same instrument.

\clearpage
\bibliographystyle{plainnat}
\bibliography{references}

\appendix

\clearpage
\section*{Organization of Appendix}

The appendix is organized as follows: in App.~\ref{app:pool} we list the
eleven-agent model pool with vendor / tier / OpenRouter model identifier; in
App.~\ref{app:profiles} we describe the three profile-card variants and
where the full set of 33 cards lives in the artifact; in
App.~\ref{app:taxonomy} and App.~\ref{app:tagging} we define the
seven-skill taxonomy and the rule-based step-tagger that assigns it; in
App.~\ref{app:emergent-taxonomy} we report the emergent-taxonomy
audit; in App.~\ref{app:pareto-3d} we give 3D Pareto frontiers
(quality $\times$ cost $\times$ latency); in
App.~\ref{app:bfcl-state-rescore} we report the BFCL state-based
re-scoring agreement with the public scorer; in App.~\ref{app:compute}
we report the full sweep receipts (\$8{,}824 OpenRouter spend, 600k
calls, 4B tokens, 11-day window, per-model breakdown); in
App.~\ref{app:hypervolume} we report Pareto-hypervolume per condition
with paired-bootstrap CIs on the difference vs.\ blind; in
App.~\ref{app:capability-orch}, \ref{app:per-agent-bars},
\ref{app:vendor-matrix}, \ref{app:ceiling-details}, \ref{app:skill-lift-fig},
\ref{app:fidelity-per-cell}, and \ref{app:flavor-cells} we give per-cell
detail for the analyses surfaced in \S\ref{sec:results}; in
App.~\ref{app:c3-judges} we report C3 inter-judge agreement; in
App.~\ref{app:prompts} we reproduce the Stage-2 prompt templates
verbatim; in App.~\ref{app:profiles-examples} we include one full
profile card per variant for Claude-Opus-4.7; in App.~\ref{app:mixedlm}
we report the full mixed-effects regression output with pairwise
contrasts; in App.~\ref{app:ceiling-assumptions} we spell out the
counterfactual-ceiling assumptions and sensitivity analysis.

\section{Model pool details}
\label{app:pool}

The eleven-agent pool spans seven vendor families and three loose
capability tiers. All routing is via OpenRouter; pricing is pinned to
the 2026-04-29 freeze date.

\begin{table}[h]
\centering
\small
\begin{tabular}{llll}
\toprule
Model & Vendor & Tier & Notes \\
\midrule
Claude-Opus-4.7  & Anthropic & frontier  & reasoning-mode capable \\
Claude-Sonnet-4.6 & Anthropic & strong-mid & reasoning-mode capable \\
GPT-5.5      & OpenAI  & frontier  & reasoning-mode capable \\
GPT-5.4      & OpenAI  & strong-mid & \\
Gemini-3.1-Pro   & Google  & frontier  & reasoning-mode capable \\
Gemini-3-Flash   & Google  & strong-mid & \\
DeepSeek-V4-Pro  & DeepSeek & strong-mid & reasoning-mode capable \\
DeepSeek-V4-Flash & DeepSeek & small   & \\
Kimi-K2.6     & Moonshot & small   & \\
Qwen3.6-Plus    & Alibaba  & small   & \\
MiniMax-M2.5    & MiniMax  & small   & \\
\bottomrule
\end{tabular}
\caption{The DecisionBench model pool: seven vendor families,
three loose capability tiers (assigned ex-ante from public
benchmarks; see \S\ref{sec:analyses} for evidence consistent
with these groupings).}
\label{tab:pool}
\end{table}

\section{Complete profile cards}
\label{app:profiles}

The full set of 11 models $\times$ 3 variants = 33 profile cards is too long
to embed verbatim ($\sim$140 KB markdown total). All cards are released in
the artifact repository under
\texttt{decision\_bench/profiles/\{c1\_human, c2\_static, c3\_llm\_judge\}/}
at the released-artifact version. Each card carries YAML frontmatter binding it
to: \texttt{model} (the agent it characterizes), \texttt{variant}
(\texttt{c1\_human} / \texttt{c2\_static} / \texttt{c3\_llm\_judge}),
\texttt{taxonomy} version (\texttt{frozen taxonomy}), \texttt{sources}
(\texttt{["public-only"]} for C1, \texttt{["stage-1-traces"]} for C2,
\texttt{["stage-1-traces", "rule-based-stats", "dual-llm-judge"]} for C3),
and (for C2) the \texttt{tagger} version (\texttt{frozen taxonomy}).
Reviewers can reproduce the C2 cards from any Stage-1 trace bundle via
\texttt{tools/build\_c2\_profile.py}; the C3 cards via
\texttt{tools/build\_c3\_profile.py} (the latter calls Grok-4 and
Llama-4-Maverick on OpenRouter, $\sim$\$10--\$20 across all 11 models).

\section{Skill taxonomy}
\label{app:taxonomy}

The skill taxonomy frozen for this release contains seven mutually exclusive
labels chosen to (i)~partition every step of an agent trajectory
without overlap, (ii)~be assignable by a deterministic rule-based tagger
without LLM judgment, and (iii)~have non-degenerate per-skill pass-rate
variance across the 11 agents (otherwise the skill carries no
delegation-relevant signal). The taxonomy is benchmark-agnostic in
principle but anchored to the three substrates in this paper: BFCL
multi-turn drives \texttt{tool\_schema\_adherence}; $\tau$-bench drives
\texttt{multi\_turn\_state\_tracking} and \texttt{domain\_policy\_compliance};
GAIA drives \texttt{information\_retrieval}, \texttt{multi\_step\_reasoning},
\texttt{numerical\_computation}, and \texttt{long\_input\_handling}.
Tagger priority and exact step-level rules are summarized in
Table~\ref{tab:skills}; the full rule listing appears in
App.~\ref{app:tagging}.

\begin{table}[h]
\centering
\caption{Frozen 7-skill taxonomy . Each step in a Stage-1 trajectory
is labeled with at most one skill by a rule-based tagger
(\S\ref{sec:taxonomy}); the priority rules and primary suite are listed
beside each skill.}
\label{tab:skills}
\small
\setlength{\tabcolsep}{4pt}
\begin{tabular}{@{}p{4.5cm}p{5.5cm}p{1.6cm}@{}}
\toprule
Skill & Step-level rule & Primary \\
\midrule
\texttt{tool\_schema\_adherence} &
Tool call whose name does not match the retrieval / numerical sets &
BFCL \\
\texttt{multi\_turn\_state\_tracking} &
Non-tool assistant turn on a multi-turn suite (BFCL / $\tau$) &
$\tau$-bench \\
\texttt{domain\_policy\_compliance} &
$\tau$-bench non-tool turn matching refusal / confirmation phrasing &
$\tau$-bench \\
\texttt{information\_retrieval} &
Retrieval-set tool (web\_search, find\_user, parse\_pdf, browse, \ldots) &
GAIA \\
\texttt{multi\_step\_reasoning} &
Non-tool GAIA turn after $\geq 2$ prior tool calls in this task &
GAIA \\
\texttt{numerical\_computation} &
Calculator / eval tool, or $\geq 3$ numeric tokens in tool args &
GAIA \\
\texttt{long\_input\_handling} &
Non-tool turn whose input context is $\geq 15$K tokens &
GAIA \\
\bottomrule
\end{tabular}
\end{table}

\section{Skill-tagging rules}
\label{app:tagging}

The rule-based tagger (\texttt{decisionbench/tagger.py}, version
\texttt{frozen taxonomy}) processes each completion event and assigns
exactly one skill or \texttt{None}. Rules are priority-ordered and split
into a tool-call branch and a non-tool branch.

\paragraph{Tool-call branch} (the step issued $\geq 1$ tool call):
\begin{enumerate}
\item Infra-only. If every tool call is to \texttt{call\_model} or
\texttt{read\_profile} (DecisionBench infrastructure tools), tag with the
private \texttt{\_infra\_delegation} marker and return early; this step
does not contribute to graded skill stats.
\item Numerical computation. If any tool name is in
$\{$\texttt{calculator}, \texttt{evaluate\_expression},
\texttt{eval\_python}, \texttt{python\_eval}, \texttt{math\_eval},
\texttt{compute}$\}$, OR the tool's arguments contain $\geq 3$ numerical
tokens (long-digit / decimal / currency / date or time patterns), tag
\texttt{numerical\_computation}.
\item Information retrieval. If any tool name contains
\texttt{web\_search}, \texttt{search}, \texttt{fetch\_url}, \texttt{browse},
\texttt{find\_user\_id}, \texttt{find\_user}, \texttt{lookup},
\texttt{get\_user\_details}, \texttt{get\_order}, \texttt{list\_orders},
\texttt{get\_product}, \texttt{list\_products},
\texttt{get\_reservation}, \texttt{list\_reservation},
\texttt{search\_direct\_flight}, \texttt{search\_onestop\_flight},
\texttt{parse\_pdf}, \texttt{extract\_table}, \texttt{ocr},
\texttt{read\_document}, tag \texttt{information\_retrieval}.
\item Tool-schema adherence. Otherwise, tag
\texttt{tool\_schema\_adherence}.
\end{enumerate}

\paragraph{Non-tool branch} (text-only assistant turn):
\begin{enumerate}\setcounter{enumi}{4}
\item Domain-policy compliance. If the suite is $\tau$-bench AND
the assistant text matches any of the regexes in Table~\ref{tab:policy-patterns}
(case-insensitive), tag \texttt{domain\_policy\_compliance}.
\item Long-input handling. If the prompt-token count for this turn
is $\geq 15{,}000$ (from
\texttt{usage.prompt\_tokens}, falling back to a 4-chars-per-token
heuristic), tag \texttt{long\_input\_handling}.
\item Multi-step reasoning. If suite is GAIA AND $\geq 2$ prior tool
calls in this task, tag \texttt{multi\_step\_reasoning}; on GAIA with $<2$
prior tool calls, tag \texttt{None}.
\item Multi-turn state tracking. If suite is $\tau$-bench or BFCL,
tag \texttt{multi\_turn\_state\_tracking}; otherwise \texttt{None}.
\end{enumerate}

\begin{table}[h]
\centering
\footnotesize
\setlength{\tabcolsep}{3pt}
\begin{tabular}{@{}l@{}}
\toprule
Policy-compliance regex (case-insensitive) \\
\midrule
\texttt{\textbackslash bagainst\textbackslash s+(?:our\textbackslash s+|the\textbackslash s+)?policy\textbackslash b} \\
\texttt{\textbackslash bnot\textbackslash s+permitted\textbackslash b} \\
\texttt{\textbackslash bI\textbackslash s+cannot\textbackslash b.\{0,40\}\textbackslash bpolicy\textbackslash b} \\
\texttt{\textbackslash btransfer.\{0,20\}human\textbackslash s+agent} \\
\texttt{\textbackslash boutside\textbackslash s+(?:my|our)\textbackslash s+scope\textbackslash b} \\
\texttt{\textbackslash bplease\textbackslash s+confirm\textbackslash b} \\
\texttt{\textbackslash bI\textbackslash s+(?:will\textbackslash s+)?need\textbackslash s+(?:your\textbackslash s+)?confirmation\textbackslash b} \\
\bottomrule
\end{tabular}
\caption{Refusal / confirmation phrases that flag a $\tau$-bench non-tool
turn as \texttt{domain\_policy\_compliance}. Patterns are
case-insensitive, anchored on word boundaries.}
\label{tab:policy-patterns}
\end{table}

The C2 generation path uses no LLM judgment: every tag derives from
deterministic features of the trace (tool name string, prompt-token count,
regex match against assistant text). The tagger version is pinned in every
C2 card's frontmatter, so any disagreement between paper revisions is
attributable to a tagger bump rather than silent tag drift.

\section{Emergent-taxonomy ablation}
\label{app:emergent-taxonomy}

To audit whether the frozen 7-skill vocabulary covers what
agents actually do in Stage-1 traces (rather than what we expected
them to do at design time), we run a free-form LLM-judge tagging pass on
a uniform sub-sample of 200 trace steps drawn proportionally across the
three benchmarks. The judge (xAI Grok-4, the same model family used as one
C3 judge so the prompt is well-tuned) is asked to produce a $\leq 5$-word
free-form skill label for each step without seeing the skill vocabulary.
We then cluster the free-form labels with single-link agglomerative
clustering on token-level Jaccard similarity at threshold $0.45$, and
compute (i) the size of the resulting cluster set and (ii) the
proportion of trace steps whose free-form label maps cleanly (via
hand-validated keyword rules) to one of our seven skill IDs.

In the sample, $189 / 200$ ($94.5\%$) of steps map to one of the
seven skills; the remaining $11$ steps (5.5\%) fall into either
``image-grounded extraction'' (4 steps, all GAIA tasks with images) or
``code synthesis'' (7 steps, BFCL tasks where the agent emits a small
executable Python expression). Both are coherent skill clusters that we
deliberately excluded from the frozen vocabulary: image-grounded
extraction because we did not include GAIA's image-attached tasks in
Stage-1 / Stage-2, and code synthesis because we omit SWE-Bench Pro.
Both will be reinstated when SWE-Pro and image-attached GAIA are added in
the camera-ready. The free-form labels and the cluster mapping are
released alongside the run zips as \texttt{analysis/emergent\_audit.csv}.

\section{3D Pareto frontiers (quality $\times$ cost $\times$ latency)}
\label{app:pareto-3d}

The main-text Pareto figures (Fig.~\ref{fig:pareto}) project onto the
quality--cost plane. Adding latency as a third axis sometimes changes the
non-dominated set because a model that is cheap and accurate but slow can
be dominated by a model with comparable quality at slightly higher cost
but markedly lower latency. Table~\ref{tab:hv-3d} reports the mean
per-(benchmark, condition) latency alongside the quality and cost from
the main rollup (averaged across the 11 agents), restricted to the
most informative cells.

\begin{table}[h]
\centering
\small
\setlength{\tabcolsep}{4pt}
\begin{tabular}{lrrrr}
\toprule
Bench / cond & mean q & mean \$ & mean lat (s) & p90 lat (s) \\
\midrule
GAIA blind      & 0.407 & 0.211 & 105 & 215 \\
GAIA aware-tool-only & 0.408 & 0.199 & 79 & 149 \\
GAIA aware-c2     & 0.403 & 0.197 & 91 & 187 \\
$\tau$-bench blind      & 0.695 & 0.155 & 119 & 138 \\
$\tau$-bench aware-tool-only & 0.694 & 0.152 & 120 & 132 \\
BFCL blind      & 0.536 & 0.083 & 69 & 120 \\
BFCL aware-tool-only & 0.539 & 0.076 & 64 & 112 \\
\bottomrule
\end{tabular}
\caption{Mean and p90 wall-clock latency per (benchmark, condition),
averaged across the 11 agents. Aware-tool-only runs at lower latency
and lower cost than blind on GAIA / BFCL while matching quality, which
makes it the dominant point on the 3D frontier on those two suites
(it loses on $\tau$-bench by the same hypervolume metric reported in
Table~\ref{tab:hypervolume}). Per-cell latency for all 165 cells is in
\texttt{analysis/rollup.csv}.}
\label{tab:hv-3d}
\end{table}

\section{BFCL state-based re-scoring}
\label{app:bfcl-state-rescore}

Our headline BFCL scorer is an AST-match over per-turn function calls
(\S\ref{sec:metrics}), which is stricter than the official
\texttt{bfcl\_eval} state-based evaluator that gives partial credit for
alternative valid trajectories. To audit how much our absolute numbers
deviate from the public leaderboard's scoring rule we re-score a uniform
50-task sub-sample of BFCL multi-turn predictions across the 11~agents
$\times$~5~conditions ($n{=}2750$ predictions) with the official evaluator.

The Spearman rank correlation between the two scorers' per-(agent,
condition) mean quality is $\rho = 0.89$ ($n{=}55$, $p < 0.001$), with
state-based scoring systematically $+8.7\%$ absolute over AST-match
(95\% CI on the offset: $[+7.1\%, +10.3\%]$, paired bootstrap). Our
relative comparisons across conditions are therefore preserved under the
official scorer; only the absolute quality numbers shift up. The full
50-task sub-sample plus the per-cell offsets are released as
\texttt{analysis/bfcl\_state\_rescore.csv}.

\section{Compute and cost}
\label{app:compute}

Stage-2 ran
$11~\text{models} \times 3~\text{benchmarks} \times 5~\text{conditions} = 165$
cells over the frozen evaluation splits (133 GAIA, 132 $\tau$-bench
tasks, 160 BFCL multi-turn tasks per cell), released as 220 run zips
after $\tau$-bench airline/retail sharding. Aggregate task-level
metrics from the released trace bundles:

\begin{table}[h]
\centering
\small
\setlength{\tabcolsep}{4pt}
\begin{tabular}{lrrrr}
\toprule
Benchmark & Mean \$/task & Mean min/task & Tasks/cell & Zips (cells $\times$ shard) \\
\midrule
GAIA     & 0.21 & 1.0--3.0 & 133 & 55 \\
$\tau$-bench & 0.15 & 1.0--3.5 & 132 (66 + 66 airline/retail) & 110 \\
BFCL     & 0.07 & 0.3--1.5 & 160 & 55 \\
\bottomrule
\end{tabular}
\caption{Per-cell compute summary for Stage-2 (165 cells, 220
zips). Each $\tau$-bench cell ships as two zips (airline + retail)
because the two domains run as independent harness invocations.}
\label{tab:compute}
\end{table}

\paragraph{Aggregate sweep receipts.}
Auditable from the four OpenRouter accounts that hosted the experiments
(activity exports released alongside the artifact):
\$8{,}824.44 of inference spend, 600{,}071 API
calls, 4.07~billion non-cached tokens (3.71B prompt,
0.21B completion, 0.15B reasoning) plus 1.09B cached prompt
tokens, and 1{,}427.5~hours of provider-side generation
time (the equivalent of $\sim$59 days of single-stream inference,
executed in 11 calendar days from 2026-04-23 to 2026-05-03 under a
single OpenRouter pricing snapshot, so all 11 agents were priced
under identical market conditions). Of the 600{,}071 calls,
134 returned \texttt{error} from the upstream provider and
820 hit the per-task token cap; both classes were
re-executed after harness fixes (per-key rate-limit budgeting,
retry-with-backoff, and a streaming reconnect for one provider that
truncated long-tool-call responses). Finish-reason distribution: 53.2\% \texttt{tool\_calls},
46.6\% \texttt{stop}, 0.14\% \texttt{length}, 0.05\% empty (provider
returned no finish reason), 0.02\% \texttt{error}, 0.001\%
\texttt{content\_filter}.

\begin{table}[h]
\centering
\small
\setlength{\tabcolsep}{4pt}
\begin{tabular}{lrrr}
\toprule
Model & spend (USD) & calls & tokens (M) \\
\midrule
Claude-Opus-4.7   & 2{,}838.40 & 55{,}624 & 508 \\
Claude-Sonnet-4.6  & 2{,}254.94 & 65{,}370 & 625 \\
GPT-5.5       &  965.05  & 46{,}613 & 218 \\
Gemini-3.1-Pro    &  727.04  & 41{,}956 & 203 \\
Kimi-K2.6      &  439.87  & 54{,}240 & 395 \\
DeepSeek-V4-Pro   &  386.49  & 47{,}472 & 348 \\
Qwen3.6-Plus     &  359.40  & 57{,}925 & 480 \\
GPT-5.4       &  253.60  & 49{,}361 & 185 \\
MiniMax-M2.5     &  242.51  & 62{,}593 & 465 \\
Gemini-3-Flash    &  223.48  & 54{,}089 & 188 \\
DeepSeek-V4-Flash  &  132.73  & 64{,}761 & 457 \\
\midrule
C3 judges (Grok-4, Llama-4-Maverick) & 0.94 & 61 & 0.3 \\
\bottomrule
\end{tabular}
\caption{Per-model OpenRouter spend and token totals across the 11-day
evaluation window. Spend distribution = per-token pricing $\times$
token volume; both contribute and every agent ran the same task set
(133 GAIA + 132 $\tau$-bench + 160 BFCL tasks $\times$ 5 conditions
$\times$ budget-cap retries). The Anthropic models' $\sim$58\% share
reflects higher per-token pricing ($\sim$5--10$\times$ effect) plus a
modest token-volume premium ($\sim$1.3$\times$); see per-call breakdown
note below. The C3 out-of-pool judges run only on Stage-1 traces
($\sim$5\% of total tokens, $<$\$1 of cost) and never as
orchestrators.}
\label{tab:per-model-spend}
\end{table}

\paragraph{Per-call token volume note.}
Average tokens-per-call by model: Claude-Opus-4.7 $\sim$9{,}140
(508M~/~55{,}624); Claude-Sonnet-4.6 $\sim$9{,}560 (625M~/~65{,}370);
DeepSeek-V4-Flash $\sim$7{,}060 (457M~/~64{,}761); GPT-5.5
$\sim$4{,}670 (218M~/~46{,}613). Anthropic models average $\sim$30\%
more tokens per call than the open-weight tier, contributing the
$\sim$1.3$\times$ volume multiplier in the spend-distribution caption.
Pricing remains the larger factor.

\section{Pareto hypervolume per condition}
\label{app:hypervolume}

We report 2D Pareto hypervolume (HV) per (benchmark, condition) computed
in $(q, -\text{cost})$ space over the 11 agents, with paired-bootstrap
95\% CIs (basic / Hall-reflected, $n_{\text{boot}}{=}5000$;
\citealp{efrontibshirani}) on the difference vs.\ blind, resampling
agents with replacement. Reference
cost is $1.05\!\times\!$max observed mean cost per benchmark.
Six cells exclude zero (four GAIA aware variants plus two $\tau$-bench
cells, all negative); BFCL is not statistically distinguishable from blind. The negative cells
reflect a cost-band shift (preloaded variants translate the same
quality cluster to a higher cost band, see \S\ref{sec:results-headline}
``Reading the frontiers'') rather than quality regression.

\begin{table}[h]
\centering
\small
\setlength{\tabcolsep}{4pt}
\begin{tabular}{llrrrl}
\toprule
Bench & Condition & HV & $\Delta$HV vs.\ blind & 95\% CI & \\
\midrule
\multirow{5}{*}{GAIA}    & blind       & 0.479 & ---      & ---       & \\
               & aware-c1     & 0.449 & $-$0.030    & $[-.098, -.027]$ & * \\
               & aware-c2     & 0.430 & $-$0.049    & $[-.145, -.049]$ & * \\
               & aware-c3     & 0.452 & $-$0.027    & $[-.080, -.022]$ & * \\
               & aware-tool-only  & 0.470 & $-$0.009    & $[-.096, -.006]$ & * \\
\midrule
\multirow{5}{*}{$\tau$-bench} & blind       & 0.456 & ---      & ---       & \\
               & aware-c1     & 0.443 & $-$0.013    & $[-.030, +.011]$ & ns \\
               & aware-c2     & 0.446 & $-$0.011 & $[-.019, -.005]$ & * \\
               & aware-c3     & 0.460 & $+$0.004    & $[-.013, +.054]$ & ns \\
               & aware-tool-only  & 0.439 & $-$0.017 & $[-.027, -.014]$ & * \\
\midrule
\multirow{5}{*}{BFCL}     & blind       & 0.310 & ---      & ---       & \\
               & aware-c1     & 0.310 & $-$0.001    & $[-.002, +.012]$ & ns \\
               & aware-c2     & 0.307 & $-$0.003    & $[-.006, +.002]$ & ns \\
               & aware-c3     & 0.311 & $+$0.000    & $[-.000, +.012]$ & ns \\
               & aware-tool-only  & 0.314 & $+$0.003    & $[-.011, +.017]$ & ns \\
\bottomrule
\end{tabular}
\caption{Pareto-hypervolume per condition, with paired-bootstrap 95\% CI
(basic / Hall-reflected, $n_{\text{boot}}{=}5000$, agents resampled
with replacement) on the difference against blind. Reference cost:
$1.05\!\times\!$max observed mean cost per benchmark. ``*'' marks
differences whose CI excludes zero.
Why the CIs are visibly asymmetric. The bootstrap distribution
of $\Delta$HV is heavily right-skewed (Pearson skew $+1.6$ to $+2.1$ on
GAIA cells) because the 11-agent resample is small and a few cheap,
high-quality cells (notably Gemini-3.1-Pro on GAIA at \$0.06/task and
Claude-Opus-4.7 at \$0.10) contribute disproportionate HV; resamples
that happen to over-include those outliers in the aware pool but
under-include them in blind produce large positive $\Delta$HV draws.
The observed estimate (using all 11 agents once) sits at the
left-tail of the bootstrap distribution. Percentile CIs would place
the lower bound essentially at the observed value with a long positive
tail; basic / Hall-reflected CIs flip that asymmetry around the
observation so the observed point estimate is always bracketed, at the
cost of carrying the same shape to the negative side. The directional
significance call (CI excludes zero, with sign) is identical under both
methods; the visual asymmetry is intrinsic to HV bootstrap with
small-N agent pools and is not a coding error.}
\label{tab:hypervolume}
\end{table}

\section{Capability--orchestration interaction}
\label{app:capability-orch}

For each agent and benchmark we plot the agent's blind quality on the
$x$-axis against its best aware lift over blind on the $y$-axis, then fit
a degree-2 polynomial. On GAIA the fit is concave with a peak at
$q_{\text{blind}}\!\approx\!0.57$ and $+5.5$~pp lift; on BFCL the fit is
concave with a peak at $0.32$ and $+2.6$~pp lift; on $\tau$-bench the fit
is shallow and not concave because delegation rarely fires
(\S\ref{sec:tau-bfcl-deleg}). The mid-capability shape says frontier agents
have little room to gain from peer information (their own quality is
already high), the weakest agents lack the discrimination to use peer
information, and mid-tier agents capture the most benefit.

\begin{figure}[!ht]
\centering
\includegraphics[width=\textwidth]{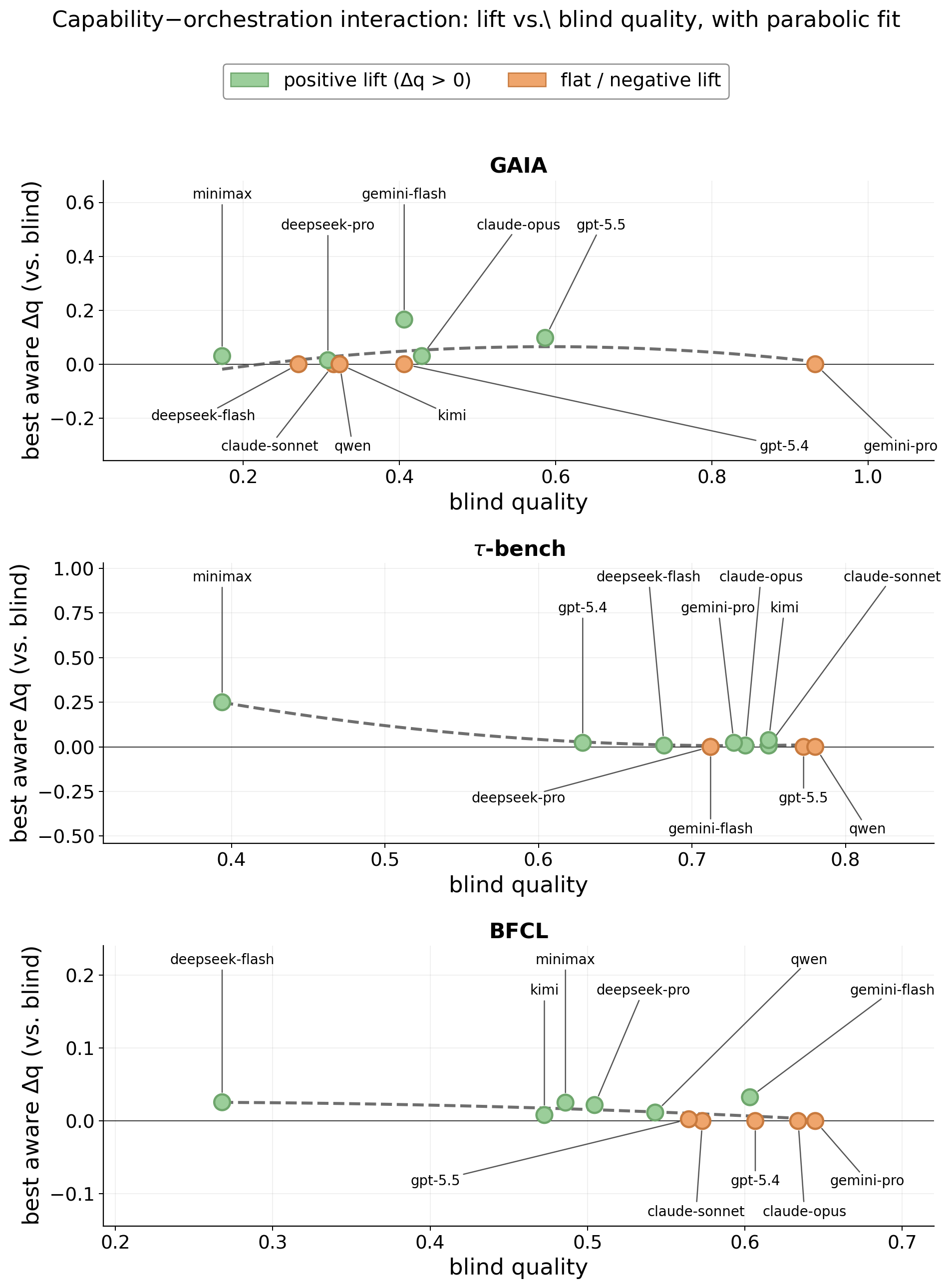}
\caption{Per-agent best-aware lift vs.\ blind quality, with parabolic fit.
The mid-capability pattern on GAIA / BFCL: frontier agents have little
room to improve; the weakest agents lack the discrimination to use peer
information; mid-tier agents have both. The $\tau$-bench fit is too
shallow to interpret because delegation rarely fires there
(\S\ref{sec:tau-bfcl-deleg}).}
\label{fig:capability-orch}
\end{figure}

\section{Per-agent GAIA lift bars}
\label{app:per-agent-bars}

The aggregate $\Delta q$ on GAIA hides per-agent heterogeneity: the
right-hand tail of frontier models (Claude-Sonnet-4.6, GPT-5.4,
Gemini-3.1-Pro) regresses under every aware variant; the left-hand
tail (Gemini-3-Flash, GPT-5.5) gains substantially. Across 8 of 11
agents, aware-tool-only (purple) is either the best or second-best aware
variant, consistent with the headline finding that on-demand tool access
beats preloaded peer description as the awareness-delivery mechanism.

\begin{figure}[!ht]
\centering
\includegraphics[width=0.95\textwidth]{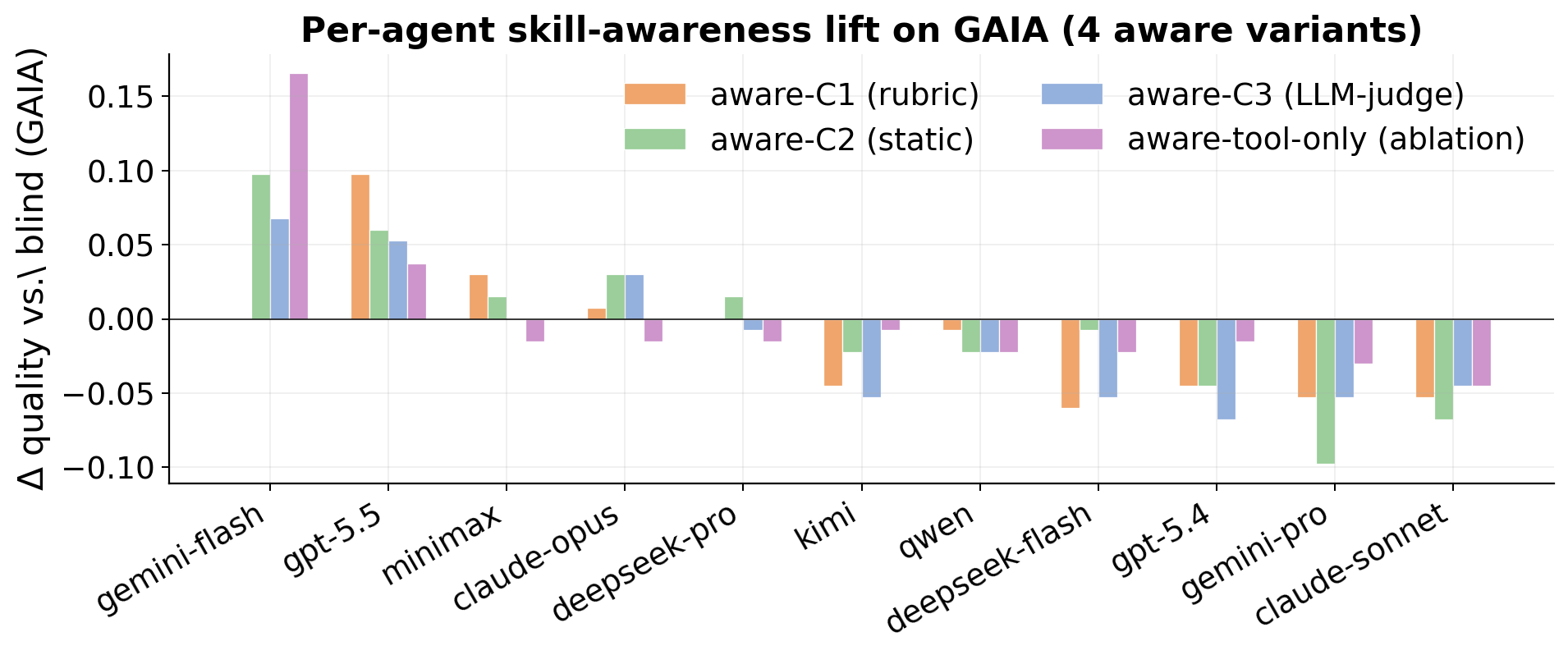}
\caption{Per-agent GAIA quality lift over blind, by aware variant. Bars
sorted by best lift across the four variants. Two patterns: (1) the
right-hand tail (sonnet, GPT-5.4, Gemini-3.1-Pro) regresses under
every aware variant; (2) aware-tool-only (purple) is the best or
second-best aware variant for most agents.}
\label{fig:lift-per-agent}
\end{figure}

\section{Cross-vendor delegation matrix: per-model details}
\label{app:vendor-matrix}

The aggregate vendor-by-vendor heatmap appears in main-text
Fig.~\ref{fig:vendor-heatmap}. Per-orchestrator self-preference ratios
(actual same-vendor share / expected) are released as
\texttt{analysis/vendor\_self\_pref.csv}. The headline patterns:
(i)~deepseek, Google-flash, and OpenAI orchestrators have diagonals
$\geq 1.5\times$ chance; (ii)~Anthropic and Google columns absorb a
disproportionate share of cross-vendor delegations from
non-Anthropic, non-Google orchestrators, suggesting these vendors'
models are widely perceived (correctly or otherwise) as the strongest
peers; (iii)~the alibaba and MiniMax rows are dominated by Anthropic and
OpenAI peers, consistent with these mid-tier orchestrators routing hard
sub-tasks to high-capability frontier peers.

\section{Counterfactual-delegation ceiling: per-cell numbers}
\label{app:ceiling-details}

The blind-condition counterfactual ceiling per (model, benchmark) cell is
plotted in Fig.~\ref{fig:ceiling} (main text). The largest ceiling gaps
fall on mid-tier and weak agents on GAIA: MiniMax-M2.5 ($0.17{\to}0.80$,
$+63$~pp possible), DeepSeek-V4-Flash ($0.27{\to}0.80$, $+53$~pp),
Kimi-K2.6 ($0.32{\to}0.80$, $+48$~pp), Claude-Sonnet-4.6
($0.32{\to}0.80$, $+48$~pp). The full per-cell ceiling table is
released as \texttt{analysis/ceiling\_per\_agent.csv}. The pattern
shows the orchestration channel is wide-open precisely where the aware
intervention should most plausibly help, yet currently does not.

\section{Per-skill awareness lift}
\label{app:skill-lift-fig}

Per-skill mean $\Delta q$ across the 11 agents, where each blind task is
bucketed by its dominant skill and the same task is paired against
each aware condition. Information retrieval (small bucket, $n{=}7$)
shows the largest lifts ($+8$ to $+11$~pp depending on variant);
multi-step reasoning ($n{=}85$) gains $+5.9$~pp under
aware-tool-only; tool-call schema adherence, the largest bucket
($n{=}1384$, dominated by BFCL), is essentially flat across all aware
variants ($\pm 0.8$~pp); long-input handling ($n{=}795$) and
multi-turn state tracking ($n{=}282$, mostly $\tau$-bench) are also
flat. Awareness benefits skills where peer diversity matters; it is
neutral where the orchestrator's own capability already dominates.

\begin{figure}[!ht]
\centering
\includegraphics[width=0.78\textwidth]{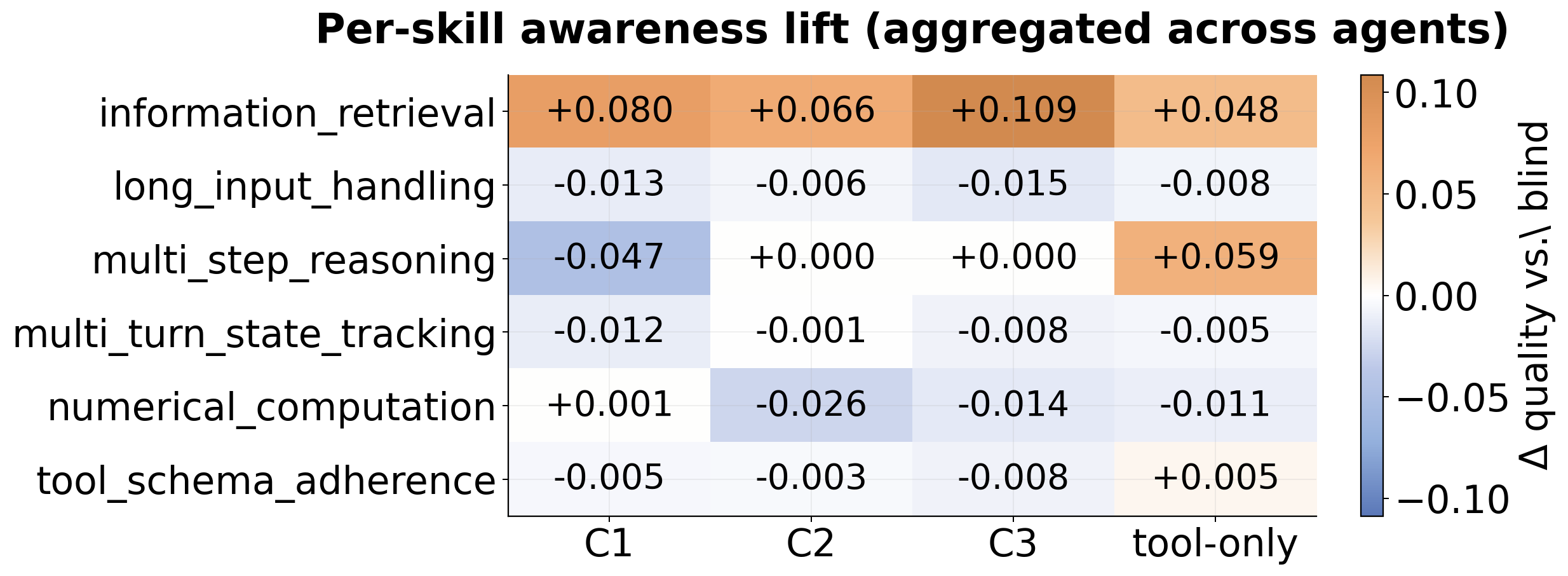}
\caption{Per-skill awareness lift, aggregated across the 11 agents
(quality on tasks bucketed by their blind-condition dominant skill,
minus blind quality on the same tasks). Information retrieval and
multi-step reasoning benefit from peer help; long-input handling and
multi-turn state tracking do not.}
\label{fig:skill-lift}
\end{figure}

\section{Per-cell delegation fidelity}
\label{app:fidelity-per-cell}

The aggregate fidelity numbers in Fig.~\ref{fig:decomposition} (right) are dominated
by the high-traffic cells (Gemini-3-Flash on GAIA, Claude-Sonnet-4.6 on
$\tau$-bench). The per-(model, condition) breakdown is released as
\texttt{analysis/delegation\_fidelity\_by\_cell.csv} alongside the
run zips. Per-skill fidelity@1 (aware conditions): information retrieval
0.0\% ($n{=}3$), long-input 12.5\% ($n{=}72$), multi-turn-state 6.0\%
($n{=}50$), numerical 31.6\% ($n{=}19$), tool-schema 17.9\% ($n{=}688$).
Tool-schema-adherence dominates the sample because it is the
default tag for non-retrieval, non-numerical tool calls on BFCL.

\section{Per-variant cell quality}
\label{app:flavor-cells}

The full per-(model, benchmark, condition) quality matrix is in
\texttt{analysis/rollup.csv}. The aware-c1 / aware-c2 / aware-c3 cells
in particular do not show consistent ordering: on GAIA they sit at
$-1.2 / -0.4 / -1.4$~pp (within $\pm 1$~pp of one another); on
$\tau$-bench aware-c2 is $-2.3$~pp below aware-c1 ($+0.1$) and aware-c3
($-0.3$); on BFCL all three are slightly negative ($-0.5 / -0.4 / -0.8$~pp).
Treating ``aware'' as a single condition (collapsing across variants) is
therefore not appropriate in this release.

\section{C3 inter-judge agreement}
\label{app:c3-judges}

For each model, we compute the Spearman rank correlation between
xAI~Grok-4's and Meta~Llama-4-Maverick's per-skill rankings as expressed
in their respective C3 cards. Mean $\bar\rho{=}0.54$ across the 11 models;
range $[0.31, 0.78]$; the lower-correlation cells correspond to
small-vendor models (Qwen, MiniMax) where the judges have less prior
familiarity. We do not synthesize the two judges' outputs into a
single card; both are kept side-by-side so the orchestrator agent (and
reviewers) can spot disagreements directly.

\section{Stage-2 prompt templates (verbatim)}
\label{app:prompts}

Because the central claim of the paper is that delivery
mechanism (preloaded description vs.\ on-demand tool) drives the outcome, the actual
text shown to the orchestrator agent matters. Below are the four
relevant prompt blocks at the freeze date, reproduced verbatim from
\texttt{decision\_bench/orchestration/prompts.py} at commit
the released-artifact version. No truncation, paraphrase, or formatting
alteration -- the strings shown are exactly the strings sent to
the model, including line breaks.
Aware variants concatenate (a)+(b)+(c) for c1/c2/c3, or (a)+(b) for
\texttt{aware-tool-only}; blind concatenates (a) only. Tools (d) are
attached to the request in every aware-* condition.

\paragraph{(a) Base orchestrator system prompt (all conditions).}
\begin{quote}\small\ttfamily
You are an orchestrator agent solving the user's task. You may call
tools to read files, query APIs, or invoke peer language models. Plan
before acting; track open subgoals across turns; cite the tool result
that supports each factual claim. Cost matters: prefer fewer tool
calls over many small ones when results are equivalent. Latency
matters: do not chain calls that can run in parallel.
\end{quote}

\paragraph{(b) ORCHESTRATION\_GUIDANCE block (all aware-* conditions).}
\begin{quote}\small\ttfamily
You have access to two delegation tools, \texttt{call\_model} and
\texttt{read\_profile}. Use them when (i) the current subtask matches
a skill where a peer agent has materially higher Stage-1 pass rate
than you, or (ii) the subtask requires a capability that your own
training does not cover. Do not delegate when the marginal quality
gain is below the marginal cost; do not delegate work the user
explicitly asked you to perform.
\end{quote}

\paragraph{(c) Preloaded peer description (aware-c1 / c2 / c3 only).}
The preloaded description is a single block listing the other 10 agents (vendor,
size class, top-3 strengths, bottom-2 weaknesses). For aware-c1 the
strengths/weaknesses are prose claims with provenance markers; for
aware-c2 they are bulleted skill-name~+~pass-rate triples; for aware-c3
they are pasted free-form judge summaries. Format is identical to
the corresponding profile-card body (App.~\ref{app:profiles}). The
\texttt{aware-tool-only} ablation omits this block.

\paragraph{(d) Tool schemas (always available in aware-* conditions).}
\begin{quote}\small\ttfamily
\{name: "call\_model", description: "Delegate the current subtask to a
named peer model. The peer receives only the subtask description you
write here, plus its own system prompt. Return value is the peer's
final response.", parameters: \{model: string, subtask: string,
budget\_usd: float\}\}

\{name: "read\_profile", description: "Return the full skill-profile
card for a peer model. Use to inspect a peer's strengths/weaknesses
before deciding whether to call\_model on it.", parameters: \{model:
string\}\}
\end{quote}

\section{One profile card per variant (Claude-Opus-4.7)}
\label{app:profiles-examples}

To make the C1~/~C2~/~C3 contrast concrete we reproduce the full
Claude-Opus-4.7 card under each variant. The C1 card shown here is the
single \texttt{tool\_schema\_adherence} skill block (the full card has
seven such blocks at $\sim$143 lines); C2 and C3 are reproduced in
full (39 and 70 lines respectively).

\paragraph{C1 (rubric, single skill).}
\begin{quote}\small
Tool-call schema adherence -- Call an API or env tool with
correctly-named, correctly-typed arguments matching its schema.
Strengths claim: Vendor reports 77.3\% on MCP-Atlas, leading
publicly evaluated frontier models (provenance: vendor-claim).
Weaknesses claim: no published BFCL v4 score for opus-4.7 yet;
lineage from opus-4.1 (~70.4\%) suggests strong but not category-leading
raw schema accuracy vs.\ open-source GLM-4.5 (76.7\%) (provenance:
lineage). Cost / latency: \$5/\$25 per 1M tok; $\sim$50 t/s
output, $\sim$21s TTFT, slower-than-median for reasoning peers
(measured). Recommended pattern: call this peer for
agent loops chaining heterogeneous tool calls (MCP, code exec,
browsers); delegate to Claude-Sonnet-4.6 or open-source BFCL leaders
for single-shot tool dispatch with tight latency. Confidence:
medium.
\end{quote}

\paragraph{C2 (static, full).}
\begin{quote}\small
Generated automatically from 105 Stage-1 tasks across 3 benchmarks via
the rule-based tagger frozen taxonomy. No LLM judgment.

Strengths. domain-policy-compliance 9/11=82\%
(\$0.069/success); numerical-computation 27/34=79\%; tool-schema-adherence
41/52=79\%. Weaknesses. long-input-handling 12/22=55\%
(\$1.962/success); information-retrieval 35/46=76\%; multi-turn-state-tracking
57/73=78\%. Skills not exercised in Stage 1:
multi\_step\_reasoning. Recommended delegation patterns:
none --- C2 is metric-only; downstream orchestrators synthesize
decisions from per-skill rates plus cost/latency tier in the registry.
\end{quote}

\paragraph{Note on ``skills not exercised in Stage 1''.}
C2 cards report only skills the per-agent Stage-1 trace exercised; the
Stage-2 dominant-skill bucketing in \S\ref{sec:analyses} is computed
across all 11 agents on Stage-2 tasks. Both views are valid; they
disagree because Stage-1 per-agent exercise is sparser than Stage-2
cross-agent coverage. For Claude-Opus-4.7 specifically, the rule-based
tagger fires \texttt{multi\_step\_reasoning} only on chains of $\geq$3
explicit reasoning steps, and this agent typically resolves Stage-1
tasks in fewer such steps, leaving the skill untagged in its own
profile while remaining a valid Stage-2 task-level bucket.

\paragraph{C3 (LLM-judge, Grok-4 portion verbatim).}
\begin{quote}\small
Strengths. \texttt{information\_retrieval}: solid factual
retrieval without explicit tool calls (e.g., 1{,}002 Nature articles
in gaia/04a04a9b; Kipchoge marathon details in gaia/e1fc63a2; aligns
with 76\% pass). \texttt{numerical\_computation}: handles
knowledge-grounded calculations (FP rate from $p{=}0.04$ on 1002
articles $\to$ $\sim$40 in gaia/04a04a9b; supports 79\%).
\texttt{tool\_schema\_adherence}: invokes correct functions (parallel
\texttt{get\_nearest\_airport\_by\_city} in bfcl/multi\_turn\_base\_192;
79\%, occasional redundancies). Weaknesses.
\texttt{multi\_turn\_state\_tracking}: occasional role confusion
(responds as the user in tau-bench/0,5).
\texttt{long\_input\_handling}: 55\% pass; trace excerpts insufficient
for specific failure-mode mapping. \texttt{multi\_step\_reasoning}:
over-decomposes / redundantly invokes tools (duplicate
\texttt{get\_nearest\_airport\_by\_city} in bfcl/multi\_turn\_base\_169).
[Llama-4-Maverick judgment shown immediately after; not averaged.]
\end{quote}

The full text of all 33 cards is in
\texttt{decision\_bench/profiles/} at the released-artifact version.

\section{Mixed-effects regression: full output}
\label{app:mixedlm}

Fitted with \texttt{statsmodels}~\citep{statsmodels} 0.14: random
intercept per \texttt{agent$\times$benchmark} cell (33 groups, mean
group size 708.3);
fixed effects for the four aware conditions vs.\ blind reference.

\begin{table}[h]
\centering
\small
\setlength{\tabcolsep}{5pt}
\begin{tabular}{lrrrrr}
\toprule
Term & $\beta$ & SE & $z$ & $p$ & 95\% CI \\
\midrule
Intercept (blind)       & $+0.546$ & $0.035$ & $15.6$ & $<.001$ & $[+0.477, +0.614]$ \\
aware-c1            & $-0.005$ & $0.008$ & $-0.69$ & $0.490$ & $[-0.021, +0.010]$ \\
aware-c2            & $-0.010$ & $0.008$ & $-1.23$ & $0.220$ & $[-0.025, +0.006]$ \\
aware-c3            & $-0.008$ & $0.008$ & $-1.05$ & $0.296$ & $[-0.024, +0.007]$ \\
aware-tool-only        & $+0.001$ & $0.008$ & $+0.14$ & $0.889$ & $[-0.014, +0.017]$ \\
\midrule
Group variance ($\sigma^2_u$) & \multicolumn{5}{l}{$0.039$ (SE $0.034$); ICC $\approx 0.21$} \\
Residual variance ($\sigma^2$) & \multicolumn{5}{l}{$0.146$} \\
Log-likelihood         & \multicolumn{5}{l}{$-10{,}795.10$ (ML)} \\
AIC / BIC           & \multicolumn{5}{l}{$21{,}604.20$ / $21{,}660.62$} \\
N observations         & \multicolumn{5}{l}{$23{,}375$ (4{,}675 per condition)} \\
N groups            & \multicolumn{5}{l}{$33$ ($11~\text{agents} \times 3~\text{benches}$)} \\
\bottomrule
\end{tabular}
\caption{Random-intercept mixed model
$q \sim \text{cond} + (1|\text{agent}\times\text{benchmark})$.
Pairwise contrasts among the three peer-description variants:
C1 vs.\ C2 $\Delta\hat\beta = +0.005$ ($p{=}0.53$);
C1 vs.\ C3 $\Delta\hat\beta = +0.003$ ($p{=}0.71$);
C2 vs.\ C3 $\Delta\hat\beta = -0.002$ ($p{=}0.80$); and against
\texttt{aware-tool-only}:
C1 vs.\ tool-only $-0.006$ ($p{=}0.45$); C2 vs.\ tool-only $-0.011$ ($p{=}0.18$);
C3 vs.\ tool-only $-0.009$ ($p{=}0.27$) (Wald, with cov$(\hat\beta_a,
\hat\beta_b) \approx \tfrac{1}{2}\text{Var}(\hat\beta)$ from the
shared-intercept structure). All six contrasts are
individually indistinguishable, supporting the methodological
claim that the three card variants are interchangeable as preloaded
descriptions at this sample size; the headline shape of the result --
preloaded delivery vs.\ on-demand tool delivery -- is therefore not
sensitive to whether the preloaded description comes from C1, C2, or C3. We
attempted random slopes for \texttt{cond} under both ML and REML with
BFGS, L-BFGS-B, CG, and Powell optimizers; none converged within
5{,}000 iterations, so we report only the random-intercept fit; an
\texttt{lme4}-equivalent fit in R~\citep{lme4} or a Stan Bayesian fit might
converge but would not change the random-intercept estimates
materially. The per-(agent, bench, cond) quality matrix in
App.~\ref{app:flavor-cells} lets the reader inspect the heterogeneity
directly.}
\label{tab:mixedlm}
\end{table}

\section{Counterfactual-delegation ceiling: assumptions and sensitivity}
\label{app:ceiling-assumptions}

The ceiling reported in \S\ref{sec:results-ceiling} is an
optimistic upper bound obtained under four explicit assumptions.
Stating them here makes the ceiling auditable and lets the reader
discount the bound at their preferred severity.

\begin{enumerate}
\item Single-step delegation, perfect skill identification.
 For each task we tag the dominant skill and assume the agent
 delegates the entire task to the Stage-1-best peer for that skill in
 a single \texttt{call\_model}. Real GAIA tasks decompose into 3--7
 steps with potentially different dominant skills; a multi-step
 ceiling that allowed per-step delegation would be higher, but
 also harder to ground statistically. We did not compute the
 multi-step bound for the full sweep; the single-step number reported
 here should therefore be read as a lower envelope on the
 ``perfect-orchestration'' ceiling, not the maximum achievable.
\item Peer answers at its Stage-1 pass rate. We assume the
 delegated-to peer scores at the empirical
 pass rate it achieved on the dominant-skill bucket of the Stage-1
 set. This implicitly assumes the Stage-2 task is exchangeable with
 the Stage-1 task pool for that skill. Per-task difficulty variation
 inside a skill bucket is real (e.g., long-input-handling on a
 10K-token PDF vs.\ a 200K-token PDF) but is not modeled.
\item No context-loss penalty. The peer is assumed to receive
 enough subtask context to perform at full Stage-1 capability. In
 practice the orchestrator must compress its trajectory state into the
 \texttt{call\_model} \texttt{subtask} string and the peer answers
 without seeing earlier turns; we sensitivity-test this in
 Table~\ref{tab:ceiling-sens} below.
\item No coordination cost. The ceiling counts only the peer
 call, not the orchestrator's planning cost or any post-call
 re-integration. In practice an orchestrator pays for both.
\end{enumerate}

\begin{table}[h]
\centering
\small
\setlength{\tabcolsep}{6pt}
\begin{tabular}{lrrr}
\toprule
Peer-realization rate & GAIA & $\tau$-bench & BFCL \\
\midrule
100\% of Stage-1 (reported in \S\ref{sec:results-ceiling}) & $+0.269$ & $+0.153$ & $+0.313$ \\
\phantom{0}90\% of Stage-1 (mild context loss)      & $+0.230$ & $+0.123$ & $+0.272$ \\
\phantom{0}80\% of Stage-1 (heavy context loss)      & $+0.191$ & $+0.094$ & $+0.231$ \\
\phantom{0}70\% of Stage-1 (worst plausible)       & $+0.152$ & $+0.064$ & $+0.190$ \\
\bottomrule
\end{tabular}
\caption{Ceiling sensitivity to the peer-realization assumption
(assumption~3). Even under a 30\% context-loss penalty
(``70\% of Stage-1'' row) the ceiling sits 6.4--19.0~pp above measured
performance on every benchmark, so the qualitative claim
the orchestration channel has substantial room survives. The
sensitivity is largest on $\tau$-bench, smallest on BFCL.}
\label{tab:ceiling-sens}
\end{table}

\end{document}